\documentclass{article}

\usepackage[nonatbib, preprint]{neurips_data_2021}

\usepackage[utf8]{inputenc} % allow utf-8 input
\usepackage[T1]{fontenc}    % use 8-bit T1 fonts
\usepackage{url}            % simple URL typesetting
\usepackage{booktabs}       % professional-quality tables
\usepackage{amsfonts}       % blackboard math symbols
\usepackage{nicefrac}       % compact symbols for 1/2, etc.
\usepackage{microtype}      % microtypography
\usepackage[dvipsnames]{xcolor}         % colors

\usepackage[pagebackref=true,breaklinks=true,colorlinks,citecolor=ForestGreen,bookmarks=false]{hyperref}
\usepackage{subfig}
\usepackage{amsmath}
\usepackage{amssymb}
\usepackage{overpic}
\usepackage{graphics}
\usepackage{rotating}
\usepackage{multirow}
\def\ie{\emph{i.e.}}
\def\eg{\emph{e.g.}}
\usepackage{wrapfig}
\usepackage[toc,page]{appendix}
\usepackage{pifont}

\newcommand{\secref}[1]{$\S$ \ref{#1}}
\newcommand{\eqnref}[1]{(Eq.~\ref{#1})}
\newcommand{\ourdataset}{\textit{SHD360}}
\newcommand{\OurTotalFrames}{37,403}
\newcommand{\OurTotalObjects}{6,268}
\newcommand{\OurTotalInstances}{16,238}
\newcommand{\OurTotalBaselines}{11}
\newcommand{\OurTotalVideos}{41}

\title{SHD360: A Benchmark Dataset for Salient Human Detection in 360° Videos}

\author{%
  Yi Zhang$^1$, Lu Zhang$^1$, Kang Wang$^2$, Wassim Hamidouche$^1$, Olivier Deforges$^1$
  \\
  $^1$Institut National des Sciences Appliquées de Rennes, France \\
  $^2$Northeast Petroleum University, Daqing, China\\
  $^1$\small\{\texttt{{yi.zhang1, Lu.Ge, Wassim.Hamidouche, Olivier.Deforges\}@insa-rennes.fr}}\\
  $^2$\small{\texttt{kangwang@stu.nepu.edu.cn}}\\
  }

\begin{document}

\maketitle

\begin{abstract}
  Salient human detection (SHD) in dynamic 360° immersive videos is of great importance for various applications such as robotics, inter-human and human-object interaction in augmented reality. However, 360° video SHD has been seldom discussed in the computer vision community due to a lack of datasets with large-scale omnidirectional videos and rich annotations. 
  To this end, we propose \ourdataset, the first 360° video SHD dataset which contains various real-life daily scenes.
  %partially borrowed from ASOD60K \cite{zhang2021asod60k}, which is so far the first 360° video salient object detection (SOD) dataset.
  Our \ourdataset~provides six-level hierarchical annotations for \OurTotalObjects~key frames uniformly sampled from \OurTotalFrames~omnidirectional video frames at 4K resolution. Specifically, each collected key frame is labeled with a super-class, a sub-class, associated attributes (e.g., geometrical distortion), bounding boxes and per-pixel object-/instance-level masks. As a result, our \ourdataset~contains totally \OurTotalInstances~salient human instances with manually annotated pixel-wise ground truth.
  Since so far there is no method proposed for 360° image/video SHD, we systematically benchmark \OurTotalBaselines~representative state-of-the-art salient object detection approaches on our \ourdataset, and explore key issues derived from extensive experimenting results.
  We hope our proposed dataset and benchmark could serve as a good starting point for advancing human-centric researches towards 360° panoramic data. Our dataset and benchmark is publicly available at \url{https://github.com/PanoAsh/SHD360}.
\end{abstract}

\section{Introduction}\label{sec:intro}

\begin{figure*}[t!]
	\centering
	\begin{overpic}[width=0.99\textwidth]{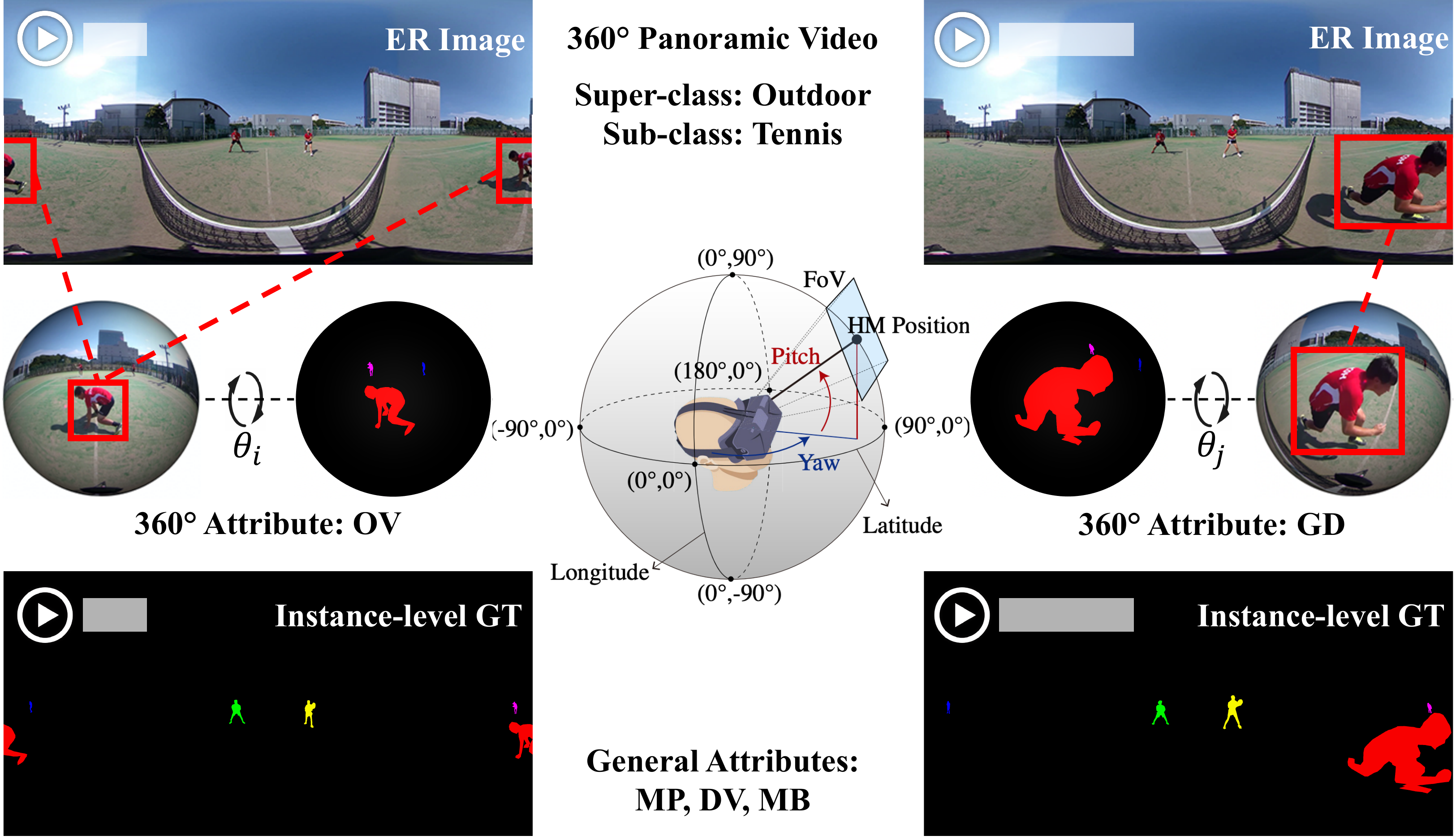}
    \end{overpic}
	\caption{\textbf{An illustration of 360° video salient human detection}. The first row, two random key frames of a 360° panoramic video from our \ourdataset. The shown 360° video frames are mapped to plane by conducting equirectangular (ER) projection. The middle row: a subject observes 360° content by moving his head to control the field-of-view (FoV) in a range of 360°$\times$180° (the figure is cited from \cite{pvshm}). The salient human instances with 360° attributes observed in spherical FoVs at specific rotation angles (\eg, $\theta_{i}$, $\theta_{j}$). The last row: corresponding annotations such as per-pixel instance-level ground truth (GT) and general attributes including MP-multiple persons, DV-distant view and MB-motion blur.}
    \label{fig:teaser}
\end{figure*}

% human-centric 360 SOD 的意义和可能用途
As the development of virtual reality (VR) and augmented reality (AR) industries, recent years have witnessed an emergence of 360° cameras, such as Facebook's Surround360, Insta360 series, Ricoh Theta and Google Jump VR, which produce 360° (omnidirectional) images capturing the scenes with a 360°$\times$180° field of view (FoV) (Figure \ref{fig:teaser}). The large-scale accessible omnidirectional images have facilitated deep learning researches towards saliency (or fixation) prediction in 360° panoramic videos (\eg, \cite{xu2018gaze,pvshm}). 360° fixation prediction, a task aims at predicting irregular regions to which subjects pay most attention when exploring 360° immersive environment, has been widely researched and applied to preliminary VR applications such as panoramic image compression and quality assessment \cite{xu2020state}. However, human-centered robotic vision and real-life AR/VR applications such as visual-language navigation, human-object interaction and inter-human interaction, all require object-level visual attention to be mimicked and/or finely predicted in real-time captured 360° immersive natural scenes. In other words, the potential algorithms are required to learn from data and segment the visual salient human entities \footnote{In this paper, we regard salient human as a specific class of general salient object categories.} in 360° panoramic scenes. Unfortunately, so far there is no dataset and benchmark for human-centric object-level saliency detection in dynamic omnidirectional scenes, which hinders the development of deep learning methodologies towards immersive human-centric applications.

% 介绍SOD和目前360 SOD的现状和不足
Salient object detection (SOD) \cite{borji2019salient,wang2021salient}, or object-level saliency detection, is an active research field in computer vision community over the past few years, with numerous exclusively-designed handcrafted and deep learning computational models. Given an 2D image, the SOD aims to pixel-wisely segment the foreground objects that grasp most of the human attention. Followed by video salient object detection (VSOD),
%\footnote{Following previous researches, we refer to salient object segmentation in 2D images/videos as SOD/VSOD.}
which appeals increasing attention from the community as the establishment of VSOD datasets (\eg, \cite{VOS,SSAV}) that collect various 2D videos and provide large-scale (Table \ref{tab:related works}) manually labeled per-pixel binary masks as ground truth. Most recently, three 360°-SOD datasets (\cite{li2019distortion,Yi2020fSOD,ma2020stage}) and one 360°-VSOD dataset, \ie, ASOD60K \cite{zhang2021asod60k}, have been proposed (see details in Table \ref{tab:related works}). %360-SSOD \cite{ma2020stage} is so far the biggest 360° SOD dataset with 1,105 images and corresponding object-level binary masks, F-360iSOD \cite{Yi2020fSOD} contains 107 images with both the object-/instance-level pixel-wise ground truth. 
All four researches emphasize unique challenges brought by 360°-SOD, for instance, frequent occurrence of small or geometrically distorted objects in equirectangular (ER) images \footnote{ER images are regarded as the most widely used lossless planar representation of 360° images.}. On the other hand, the scale of current 360°-SOD datasets are far from being satisfactory for further deep learning-based researches, when compared with SOD datasets (\eg, DUTS-TR \cite{DUTS}, as the most widely used training set, contains 10,553 images with manually labeled pixel-wise annotations). ASOD60K is a large-scale dataset however with annotations based on general object categories. 
%Furthermore, existing 360° SOD datasets \cite{li2019distortion,Yi2020fSOD,ma2020stage} simply follow annotation protocols designed for SOD in 2D domain, thus introducing bias on the varying extent.  

Considering the indispensability of VSOD datasets for immersive human-centric applications, and unsolved problems in existing 360°-SOD datasets, we thereby propose \textbf{\ourdataset}, the first large-scale 360° video dataset for salient human detection (360°-VSHD), with totally \OurTotalFrames~360° video frames reflecting various human-centric scenes (example video shown in Figure \ref{fig:teaser}). Our \ourdataset~provides diverse annotations including hierarchical categories, manually labeled per-pixel object-level and instance-level ground-truth masks, also fine-grained attributes associated with each of the video scenes to disentangle the various challenges for 360°-SOD/-VSOD/-VSHD. In a nutshell, we provide three main contributions as follows:
\begin{itemize}
    \item We propose \ourdataset, the first 360°-VSHD dataset that contains \OurTotalFrames~frames representing \OurTotalVideos~human-centric video scenes, with \OurTotalInstances/\OurTotalObjects~manually labeled per-pixel instance/object-level ground truth corresponding to \OurTotalInstances~salient human instances, and well-defined per-scene attributes to benefit model evaluation and challenge analysis.
    \item We contribute the community a comprehensive benchmark study in terms of 360°-VSHD, by systematically evaluating \OurTotalBaselines~state-of-the-art (SOTA) methods on our \ourdataset, with four widely used metrics as well as our 360° geometry-adapted S-measure.
    \item We illustrate extensive experimenting results from different perspectives, and present an in-depth analysis to highlight the key issues within the field of 360°-SOD/-VSOD/-VSHD, thus inspiring future model development.
\end{itemize}

\begin{table*}[t!]
  \centering
  \renewcommand{\arraystretch}{1}
  \setlength\tabcolsep{0.1pt}
  \resizebox{1\textwidth}{!}{
  \begin{tabular}{r|c|c|c|c|c|c}
   \toprule
   Dataset~~~~~~~~ & Task & Year & GT Scale  & GT Resolution & GT Level & Attr.
  \\
  \midrule
  ECSSD \cite{ECSSD} &  SOD & CVPR'13 & 1,000 images & max(w,h)=400, min(w,h)=139 &  obj. & 
  \\
  DUT-OMRON \cite{DUTO} & SOD & CVPR'13 & 5,168 images & max(w,h)=401, min(w,h)=139 &  obj. &
  \\
  PASCAL-S \cite{PASCALS} & SOD & CVPR'14 & 850 images & max(w,h)=500, min(w,h)=139 &  obj. &
  \\
  HKU-IS \cite{HKUIS} & SOD & CVPR'15 & 4,447 images & max(w,h)=500, min(w,h)=100 &  obj. &
  \\
  DUTS \cite{DUTS} & SOD & CVPR'17 & 15,572 images & max(w,h)=500, min(w,h)=100 &  obj. & 
  \\
  ILSO \cite{li2017instance} & SOD & CVPR'17 & 1,000 images & max(w,h)=400, min(w,h)=142 & obj.\&ins. & 
  \\
  SOC \cite{fan2018salient} & SOD & ECCV'18 & 6,000 images & max(w,h)=849, min(w,h)=161 & obj.\&ins. & \checkmark
  \\
  SIP \cite{fan2020rethinking} & SOD & TNNLS'20 & 929 images & max(w,h)=992, min(w,h)=744 & obj.\&ins. 
  \\
  \midrule
  SegTrack V2 \cite{SegV2} &  VSOD & ICCV'13 & 1,065 from 1,065 frames & max(w,h)=640, min(w,h)=212 & obj. \\
  FBMS \cite{FBMS} &  VSOD & TPAMI'14 & 720 from 13,860 frames & max(w,h)=960, min(w,h)=253 & obj.
  \\
  ViSal \cite{ViSal} &  VSOD & TIP'15 & 193 from 963 frames & max(w,h)=512, min(w,h)=240 & obj.
  \\
  DAVIS2016 \cite{DAVIS} &  VSOD & CVPR'16 & 3,455 from 3,455 frames & max(w,h)=1,920, min(w,h)=900 & obj. & \checkmark
  \\
  VOS \cite{VOS} &  VSOD & TIP'18 & 7,467 from 116,103 frames & max(w,h)=800, min(w,h)=312 & obj. & 
  \\
  DAVSOD \cite{SSAV} &  VSOD & CVPR'19 & 23,938 from 23,938 frames & max(w,h)=640, min(w,h)=360 & obj.\&ins. & \checkmark
  \\
  \midrule
  360-SOD \cite{li2019distortion} & 360°-SOD & JSTSP'19 & 500 ER images & max(w,h)=1,024, min(w,h)=512 & obj. & 
  \\
  F-360iSOD \cite{Yi2020fSOD} & 360°-SOD & ICIP'20 & 107 ER images &  max(w,h)=2,048, min(w,h)=1,024 & obj.\&ins. & 
  \\
  360-SSOD \cite{ma2020stage} & 360°-SOD & TVCG'20 & 1,105 ER images & max(w,h)=1,024, min(w,h)=546 & obj. &
  \\
  \midrule
  ASOD60K \cite{zhang2021asod60k} & 360°-VSOD & arXiv & 10,465 from 62,455 frames & max(w,h)=3,840, min(w,h)=1,920 & obj.\&ins. & \checkmark
  \\
  \midrule
  \textbf{\ourdataset} & 360°-VSHD &  arXiv & \OurTotalObjects~from \OurTotalFrames~frames & max(w,h)=3,840, min(w,h)=1,920 & obj.\&ins. & \checkmark
  \\
  \bottomrule
  \end{tabular}}
   \caption{Summary of widely used salient object detection (SOD) datasets and our \ourdataset. GT = ground truth. ER Image = equirectangular image. Attr. = attributes. obj. = object-level GT. ins. = instance-level GT. Note that all the datasets listed above provide pixel-wise annotations.}
   \label{tab:related works}
\end{table*}

\section{Related Works}\label{related_works}
\subsection{Benchmark Datasets}\label{related_benchmark}
\noindent
\textbf{360° Saliency Prediction}. As the development of commercial head-mounted displays (HMDs), 360° saliency (fixation) prediction has become a popular topic in the computer vision community, around which both the image \cite{salient360img,stfvr} and video \cite{360VHMD,vrvqa48,pvshm,vqaov,saliency360video,xu2018gaze,cheng2018cube} benchmark datasets have been proposed. Most datasets provide either head positions or eye positions of several subjects wearing HMDs and embedded eye trackers during subjective experiments. These head/eye movement records were regarded as ground truth for saliency prediction task. It is worth mentioning that, datasets such as VR-Scene \cite{xu2018gaze}, 360-Saliency \cite{saliency360video}, VQA-OV \cite{vqaov} and PVS-HMEM \cite{pvshm} all provide ground-truth eye-fixation data, thus benefiting accurate saliency prediction. However, the task of 360° fixation prediction can hardly contribute real-life VR/AR applications which require the segmentation of salient objects with finely traced boundaries.
\par
\noindent
\textbf{360°-SOD/-VSOD}. Recent researches \cite{yang2018object,zhao2020spherical,360indoorWACV2020,360sports} shift attention to object detection in 360°, however, focusing only on bounding box detection. 360°-SOD is a task to mimic human attention mechanism by pixel-wisely depicting the most significant objects from the given omnidirectional scenes. 360-SOD \cite{li2019distortion} is the first 360°-SOD dataset with 500 ER images and corresponding object-level binary masks. Followed by F-360iSOD \cite{Yi2020fSOD}, which is the first 360°-SOD dataset that provides pixel-wise instance-level ground-truth masks. 360-SSOD \cite{ma2020stage}, as the latest proposed 360°-SOD dataset, released 1,105 ER images with only object-level masks. A detailed comparison of three datasets is shown in Table \ref{tab:related works}.
%It is worth mentioning that, the above datasets simply followed salient object annotation protocols applied in 2D domain, thus introducing bias towards saliency judgements in 360° content. 
To the best of our knowledge, ASOD60K \cite{zhang2021asod60k} is the first and so far the only 360°-VSOD dataset, which provides object-/instance-level annotations of general audio-visual object classes.
\par
\noindent
\textbf{SOD and VSOD}. In 2D domain, DUTS \cite{DUTS} is so far the biggest SOD dataset, with 10,553/5019 images as training/testing set, respectively. ECSSD \cite{ECSSD}, DUT-OMRON \cite{DUTO}, PASCAL-S \cite{PASCALS} and HKU-IS \cite{HKUIS} are the most commonly applied benchmark datasets for SOD model evaluation. Besides, more recently proposed datasets such as ILSO \cite{li2017instance} and SOC \cite{fan2018salient} also provide instance-level salient object annotations. Early VSOD datasets such as SegTrack V2 \cite{SegV2}, FBMS \cite{FBMS}, ViSal \cite{ViSal} and DAVIS2016 \cite{DAVIS} (also being famous for video object segmentation) include only one or a few spatially connected salient objects in each of the annotated frames. More recent datasets such as VOS \cite{VOS} and DAVSOD \cite{SSAV} contain videos with more challenging scenes and more salient objects. Please refer to Table \ref{tab:related works} for detailed information of the widely applied SOD/VSOD datasets.
\par
\noindent
\textbf{Human Detection}. As human is the most frequent and significant participant in our daily visual data, understanding human from images and videos is of great importance in computer vision community. Popular human-centric tasks including human detection \cite{dalal2005histograms}, human re-identification \cite{varior2016gated}, human pose estimation \cite{toshev2014deeppose}, human parsing \cite{gong2017look} and human-centric relation segmentation \cite{liu2021human}. Recent large-scale human detection datasets were established for specific purposes, such as pedestrian detection (\eg, Caltech Pedestrian \cite{dollar2011pedestrian}, EuroCity Persons \cite{braun2018eurocity}) and crowd counting (\eg, UCF-CC-50 \cite{idrees2013multi}, PANDA \cite{wang2020panda}). Note that these datasets do not provide pixel-wise ground truth. To the best of our knowledge, SIP \cite{fan2020rethinking} is so far the only human-centric SOD (or, SHD) dataset which provides 929 2D images with both object-/instance-level per-pixel ground-truth masks. 

\subsection{SOD Methodologies}\label{sec:related_method}
\noindent
\textbf{SOD}. In the past few years, U-Net and feature pyramid networks (FPN) have been the most commonly used basic architectures for SOTA models (e.g., \cite{BASNet,CPD,PoolNet,EGNet,SCRN,F3Net,GCPANet,ITSD,CVPR2020LDF, MINet,SOD100K,GateNet}), which were trained on large-scale benchmark datasets (e.g., DUTS \cite{DUTS}) in a manner of fully supervision. Specifically, methods such as BASNet \cite{BASNet}, PoolNet \cite{PoolNet}, EGNet \cite{EGNet} SCRN \cite{SCRN} and LDF\cite{CVPR2020LDF} pay much attention to object boundaries detection. GateNet \cite{GateNet} was embedded with a gated module for more efficient information exchange between the encoder and decoder. With comparable detecting accuracy, methods such as CPD \cite{CPD}, ITSD \cite{ITSD} and CSNet \cite{SOD100K} focus on designing light-weighted models with significantly improved inference speed. Due to the limited space, we will not include all the SOD methods in this section (please refer to recent survey \cite{wang2021salient} for more information).
\par
\noindent
\textbf{VSOD}. Recent development of large-scale video datasets such as VOS \cite{VOS} and DAVSOD \cite{SSAV} have facilitated the development of deep learning-based VSOD. \cite{MGA}, \cite{li2018flow} and \cite{RCRNet} modeled the temporal information by combining optical flow. SSAV \cite{SSAV} mimicked human attention shift mechanism by proposing saliency-shift-aware ConvLSTM. COSNet \cite{COSNet} learned mutual features between video frames with co-attention Siamese networks. PCSA \cite{gu2020PCSA} applied self-attention to learn the relations of pair-wise frames. More recently, TENet \cite{ren2020tenet} proposed new excitation module from the perspective of curriculum learning, and achieved top performance on multiple VSOD benchmarks.
\par
\noindent
\textbf{360°-SOD}. To the best of our knowledge, DDS \cite{li2019distortion}, SW360 \cite{ma2020stage} and FANet \cite{huang2020fanet} are so far the only three models exclusively designed for 360°-SOD, which all emphasized the importance of mitigating the geometrical distortion of ER images via specific new modules. Besides, no method has been proposed for 360°-VSOD.

\begin{figure*}[t!]
	\centering
	\begin{overpic}[width=0.99\textwidth]{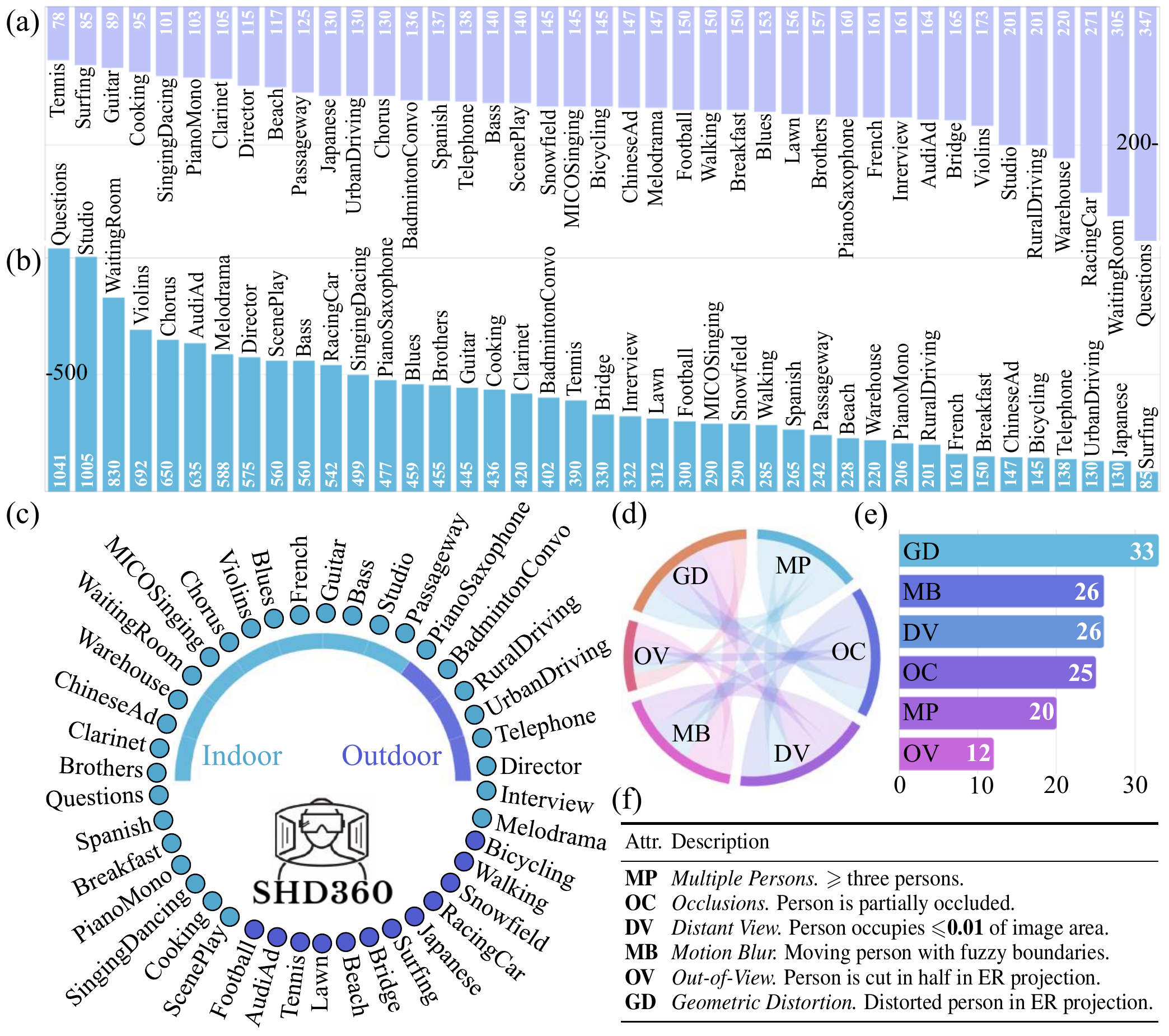}
    \end{overpic}
	\caption{Statistics of the proposed \textbf{\ourdataset}. (a)/(b) The quantity of object-/instance-level per-pixel ground-truth masks of each of the scene categories. (c) Hierarchical labels including two super-classes (indoor/outdoor) and \OurTotalVideos~scene categories. Attributes statistics including (d) and (e) which represent correlation and frequency of proposed attributes, respectively. (f) Descriptions of the six proposed attributes associated with each of the scene categories.}
    \label{fig:categories}
\end{figure*}

\begin{figure*}[t!]
	\centering
	\begin{overpic}[width=0.99\textwidth]{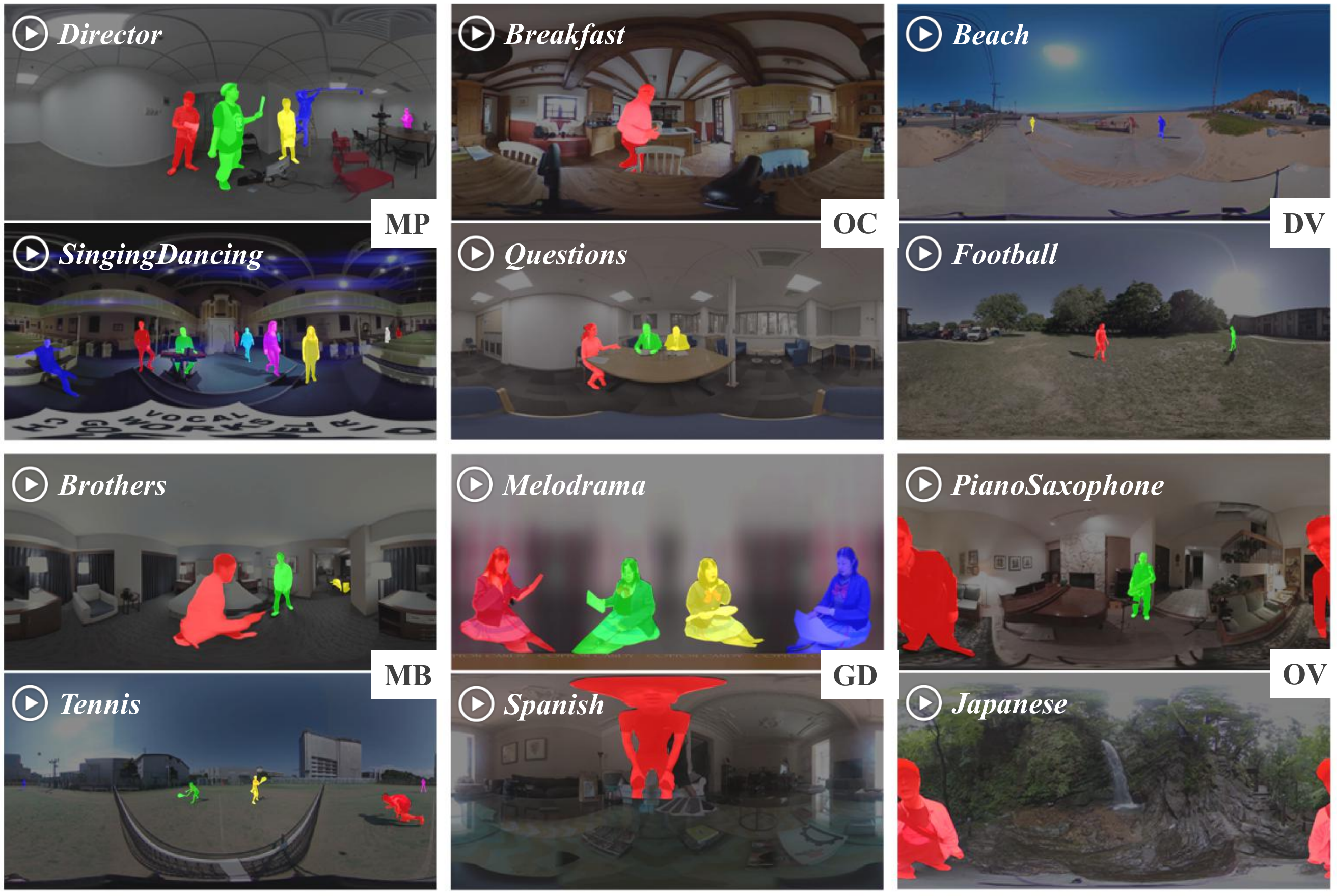}
    \end{overpic}
	\caption{Examples of instance-level pixel-wise labels and challenging attributes (please refer to Figure \ref{fig:categories} (f) for details) of our \ourdataset. Best viewed in color and zoomed in.}
    \label{fig:attributes}
\end{figure*}

\section{\ourdataset~Dataset}\label{sec:SHD360}
In this section, we introduce our \textbf{\ourdataset} from the aspects of data collection, saliency judgements, annotation pipeline and dataset statistics. Some example video frames and corresponding annotations are shown in Figure \ref{fig:attributes}. 
%Note that our \ourdataset~is the first 360° VSHD dataset.

\subsection{Video Collection}\label{sec:data_collection}
%The stimuli of our \ourdataset~are all high resolution (4K) downloads (ER videos) from YouTube, for high visual quality may facilitate future applications of our dataset based on VR/AR devices (\eg, HMDs). We first collected 101 360° panoramic videos (where 67 videos were borrowed from ASOD60K \cite{zhang2021asod60k}) representing random natural daily scenes such as concerts, sports, urban transportation, carnivals, animals, machines and architectures. Since our goal is to establish a human-centric 360°-VSOD dataset, we then carefully filtered the raw videos and further gained 69 videos centered on different human activities.
The stimuli of our \ourdataset~are from ASOD60K \cite{zhang2021asod60k}.
%
%Considering the potential ambiguity between the salient and non-salient objects in a wide FoV of 360° images, we did not simply follow the protocol applied in 2D domain. Since 2D images from popular SOD \cite{DUTS}/VSOD \cite{SSAV} datasets include scenes with limited context information based on single viewport, it is reasonable to subjectively define salient objects from usually no more than five candidate foreground objects appeared in each of the collected images. On the contrary, existing 360° SOD datasets such as 360-SOD \cite{li2019distortion} and 360-SSOD \cite{ma2020stage}, contain general natural scenes with sometimes cluttered foreground/background objects yet only annotated a few foreground objects as salient ones, based merely on pure subjective judgements of limited annotators. To avoid such an ambiguity, 
%
To serve our annotation protocol for non-ambiguous saliency judgements (please refer to \secref{sec:judging} for details), we further selected \OurTotalVideos~videos from the 69 human-centered videos, to contain only natural daily scenes with clear foreground persons involved in specific human activities. Therefore, we finally collected \OurTotalFrames~video frames representing \OurTotalVideos~indoor/outdoor panoramic human-centric scenes (Figure \ref{fig:categories} (c)), which cover multiple human activities such as singing, acting, conversation, monologue and sports.

\subsection{Saliency Judgements}\label{sec:judging}
Inspired by ViSal \cite{ViSal}, which is the first specially designed VSOD dataset with annotated salient objects according to semantic classes of videos, we annotated the salient persons based on specific human activity (Figure \ref{fig:categories} (c)) of each of the video scenes (please refer to Figure \ref{fig:attributes} and the appendices for partial/full visual examples, respectively). Specifically, since we only keep the videos with foreground persons involved in specific human activities \secref{sec:data_collection}, we considered all foreground human instances as salient targets in our \ourdataset. Compared with other widely used SOD (ECSSD \cite{ECSSD}, DUTS \cite{DUTS}, HKU-IS \cite{HKUIS}, etc.) and VSOD (DAVIS2016 \cite{DAVIS}, SegTrack V2 \cite{SegV2}, FBMS \cite{FBMS}, etc.) datasets, which contain only one or several center-biased foreground objects, our \ourdataset~is much more challenging for including multiple scattered foreground human targets (\eg,\emph{SingingDancing} shown in Figure \ref{fig:attributes}). 

\subsection{Data Annotation}\label{sec:annotation}
Generally, the hierarchical annotations of our \ourdataset~\footnote{Collecting the per-pixel labels was a laborious and time-consuming work, and it took us about three months to set up this large-scale database.} are four-fold: 1) \OurTotalVideos~scene categories respectively denote \OurTotalVideos~(12 outdoor/29 indoor) human-centric video scenes (Figure \ref{fig:categories} (c)). 2) comprehensive attributes attached to each of the collected video scenes (Figure \ref{fig:categories} (f)). 3) \OurTotalObjects~manually labeled object-level pixel-wise masks (in 4K resolution) as the ground truth for 360°-VSHD tasks. 4) \OurTotalInstances~manually labeled instance-level pixel-wise masks (in 4K resolution) corresponding to each of the salient human instances. Note that the density of object-/instance-level masks of each scene class is shown in Figure \ref{fig:categories} (a)/(b), respectively.

\subsection{Dataset Features and Statistics}\label{sec:data_attributes}
Following recent SOC \cite{fan2018salient}, DAVIS2016 \cite{DAVIS} and DAVSOD \cite{SSAV}, we further concluded six attributes to reflect the major challenges of our \ourdataset, including \emph{multiple persons (MP)}, \emph{occlusions (OC)}, \emph{distant view (DV)}, \emph{out-of-view (OV)}, \emph{motion blur (MB)} and \emph{geometrical distortion (GD)}. It is worth mentioning that, \emph{OV} and \emph{GD} were exclusively designed geometrical attributes reflected by ER images. Besides, \emph{DV} sustains stricter standard than its counterpart in 2D domain (In DAVIS2016 \cite{DAVIS}, small objects occupy $\leqslant$0.1 of the whole image area, rather than $\leqslant$0.01 as in \ourdataset), which indicates extra challenges brought by extreme small objects (far from 360° camera) when conducting 360°-SOD/-VSOD. As shown in Figure \ref{fig:categories} (d) and (e), the six proposed attributes all own high frequency and are closely related to each other, representing the challenging scenarios included in our \ourdataset~(appendices) for per-scene attributes statistics). The proposed attributes are able to support systematical quantitative evaluation of competing models (appendices), thus inspiring future model development. 

\subsection{Dataset Splits}\label{sec:data_split}
In \ourdataset, all \OurTotalVideos~videos were splitted into separate training and testing sets in the ratio of about 1:1, with a random selection strategy. Therefore, we reached a unique split consists of 21 training and 20 testing videos (3,150/3,118 key frames respectively).
%, with corresponding per-pixel instance-/object-level ground-truth.
The testing set was further divided into test-0/-1/-2 with 8/8/4 videos, respectively, according to the targets' density (maximum labeled human instances per frame). Specifically, $\leqslant$2 for test0, 3 or 4 for test1, $\geqslant$5 for test2.

\begin{table*}[t!]
  \centering
  \renewcommand{\arraystretch}{1.1}
  \setlength\tabcolsep{1.5pt}
  \resizebox{0.99\textwidth}{!}{
  \begin{tabular}{l||ccccc||ccccc||ccccc}
   \toprule
   \multirow{2}{*}{Methods} & \multicolumn{5}{c||}{Test0} & \multicolumn{5}{c||}{Test1} & \multicolumn{5}{c}{Test2}
  \\
  \cline{2-16}
   & $F_{\beta}\uparrow$ & $S_{\alpha}\uparrow$ & $E_\phi\uparrow$ & $\mathcal{M}\downarrow$ & $S_{\alpha}^{360}\uparrow$   
    & $F_{\beta}\uparrow$ & $S_{\alpha}\uparrow$ & $E_\phi\uparrow$ & $\mathcal{M}\downarrow$ 
     & $S_{\alpha}^{360}\uparrow$   
    & $F_{\beta}\uparrow$ & $S_{\alpha}\uparrow$ & $E_\phi\uparrow$ & $\mathcal{M}\downarrow$ 
     &$S_{\alpha}^{360}\uparrow$   
  \\
  \midrule 
 %  CVPR'19~CPD \cite{CPD} & .577 & .686 & .747 & .075 & .674
  CVPR'19~CPD \cite{CPD} & .529 & .686 & .682 & .075 & .674
  % & .636 & .696 & .866 & .034 & .721
    & .574 & .696 & .705 & .034 & .721
 %  & .602 & .691 & .927 & .022 & .712 \\
   & .552 & .691 & .725 & .022 & .712 \\
   
 % ICCV'19~SCRN \cite{SCRN} & {\color{green} .687} & .676 & {\color{green} .829} & .075 & .568
  ICCV'19~SCRN \cite{SCRN} & .535 & .676 & .637 & .075 & .568
  %& .689 & .721 & .924 & .032 & .653
   & .608 & .721 & .729 & .032 & .653
 % & .669 & .691 & .946 & .020 & .633 \\
  & .580 & .691 & .671 & .020 & .633 \\
  
  %AAAI'20~F3Net \cite{F3Net} & {\color{green} .687} & {\color{blue} .773} & .807 & {\color{red} .061} & {\color{red} .798}
  AAAI'20~F3Net \cite{F3Net} & {\color{blue} .677} & {\color{blue} .773} & {\color{blue} .797} & {\color{red} .061} & {\color{red} .798}
  %& .672 & .763 & .892 & .026 & {\color{red} .802}
  & .661 & .763 & {\color{green} .875} & .026 & {\color{blue} .802}
  %& .690 & .799 & .945 & .017 & {\color{green} .790} \\
  & .667 & .799 & {\color{red} .928} & .017 & .790 \\
  
  %CVPR'20~MINet \cite{MINet} & .597 & .711 & .807 & .071 & .699
  CVPR'20~MINet \cite{MINet} & .586 & .711 & .769 & .071 & .699
%  & .684 & .741 & .896 & .029 & .704
   & .669 & .741 & .835 & .029 & .704
 % & .647 & .741 & .894 & .018 & .745 \\
  & .631 & .741 & .814 & .018 & .745 \\
  
  %CVPR'20~LDF \cite{CVPR2020LDF} & .673 & .739 & .806 & {\color{green} .069} & .651
  CVPR'20~LDF \cite{CVPR2020LDF} & {\color{green} .664} & .739 & {\color{green} .786} & {\color{green} .069} & .651
 % & .724 & .769 & {\color{green} .932} & .028 & .668
  & .713 & .769 & {\color{blue} .876} & .028 & .668
 % & .661 & .766 & .891 & .017 & .657 \\
  & .648 & .766 & .850 & .017 & .657 \\
  
  %ECCV'20~CSF \cite{SOD100K} & {\color{red} .722} & {\color{red} .789} & {\color{blue} .835} & {\color{blue} .063} & {\color{blue} .738}
  ECCV'20~CSF \cite{SOD100K} & {\color{red} .703} & {\color{red} .789} & {\color{red} .814} & {\color{blue} .063} & {\color{green} .738}
 % & {\color{blue} .791} & {\color{red} .840} & {\color{blue} .936} & {\color{red} .018} & {\color{blue} .790}
  & {\color{red} .767} & {\color{red} .840} & {\color{red} .898} & {\color{red} .018} & {\color{green} .790}
 % & {\color{green} .773} & {\color{blue} .821} & {\color{green} .959} & {\color{red} .013} & .752 \\
  & {\color{green} .742} & {\color{green} .821} & .888 & {\color{red} .013} & .752 \\
  
  %ECCV'20~GateNet \cite{GateNet} & .660 & .734 & .824 & .070 & .614
  ECCV'20~GateNet \cite{GateNet} & .610 & .734 & .741 & .070 & .614
  %& .727 & {\color{green} .776} & .905 & {\color{green} .025} & .715
  & .679 & .776 & .818 & {\color{green} .025} & .715
 % & .691 & .773 & .916 & .017 & .700 \\
  & .652 & .773 & .802 & .017 & .700 \\
  
  %%TIP'21~SAMNet \cite{SAMNet} & .471 & .647 & .681 & .078 & .705
 %  TIP'21~SAMNet \cite{SAMNet} & .454 & .647 & .642 & .078 & .705
 %% & .547 & .677 & .842 & .040 & .715
 %  & .504 & .677 & .738 & .040 & .715
% % & .532 & .666 & .930 & .024 & .678\\
  % & .506 & .666 & .818 & .024 & .678\\
   
   AAAI'21~PA-KRN \cite{xu2021locate} & .596 & .695 & .727 & .073 & {\color{blue} .769} & {\color{blue} .723} & {\color{green} .787} & .855 & .028 & {\color{red} .823} & {\color{red} .769} & {\color{blue}.826} & {\color{blue} .926} & {\color{red} .013} & {\color{green} .792}  \\
  
  \midrule
  %ICCV'19~RCRNet \cite{RCRNet} & .641 & {\color{green} .740} & .808 & .070 & .680
   ICCV'19~RCRNet \cite{RCRNet} & .623 & {\color{green} .740} & .784 & .070 & .680
 % & {\color{green} .753} & {\color{blue} .821} & .913 & {\color{blue} .024} & {\color{green} .789}
  & {\color{green} .718} & {\color{blue} .821} & .873 & {\color{blue} .024} & .789
 % & .733 & {\color{red} .828} & {\color{blue} .965} & {\color{green} .016} & {\color{blue} .798} \\
  & .695 & {\color{red} .828} & {\color{green} .908} & {\color{green} .016} & {\color{blue} .798} \\
  
  %AAAI'20~PCSA \cite{gu2020PCSA} & .565 & .667 & .720 & .075 & {\color{green} .722}
  AAAI'20~PCSA \cite{gu2020PCSA} & .539 & .667 & .661 & .075 & .722
 % & .721 & .766 & .891 & .029 & .777
  & .692 & .766 & .814 & .029 & .777
  %& {\color{blue} .777} & {\color{green} .816} & .930 & {\color{blue} .015} & {\color{red} .818}\\
   & {\color{blue} .746} & .816 & .858 & {\color{blue} .015} & {\color{red} .818}\\
  
  \midrule
  % SPL'20~FANet \cite{huang2020fanet} & {\color{blue} .720} & .672 & {\color{red} .839} & .081 & .567
   SPL'20~FANet \cite{huang2020fanet} & .475 & .672 & .586 & .081 & .567
   %& {\color{red} .812} & .744 & {\color{red} .943} & .036 & .635
   & .584 & .744 & .663 & .036 & .635
 % & {\color{red} .845} & .768 & {\color{red} .981} & .021 & .660\\
    & .634 & .768 & .683 & .021 & .660\\
    
  \bottomrule
  \end{tabular}}
   \caption{Performance comparison of 8/2 SOTA SOD/VSOD methods and one 360°-SOD method over the three testing sets of our \ourdataset. $S_\alpha$ = S-measure ($\alpha$=0.5 \cite{Fan2017Smeasure}), $S_{\alpha}^{360}$ = 360° geometry-adapted S-measure, $F_\beta$
  % = maximum F-measure
   = mean F-measure 
   ($\beta^2$=0.3) \cite{Fmeasure}, $E_\phi$ = 
   %maximum E-measure 
   mean E-measure
   \cite{Fan2018Enhanced}, $\mathcal{M}$ = mean absolute error \cite{MAE}. $\uparrow$/$\downarrow$ denotes a larger/smaller value is better. The three best results of each column are in {\color{red} \textbf{red}}, {\color{blue} \textbf{blue}} and {\color{green} \textbf{green}}, respectively. 
   %Evaluation code: \url{https://github.com/zzhanghub/eval-co-sod}.
   }
   \label{tab:QuantitativeComparison}
\end{table*}

\begin{figure*}[t!]
	\centering
	\begin{overpic}[width=0.99\textwidth]{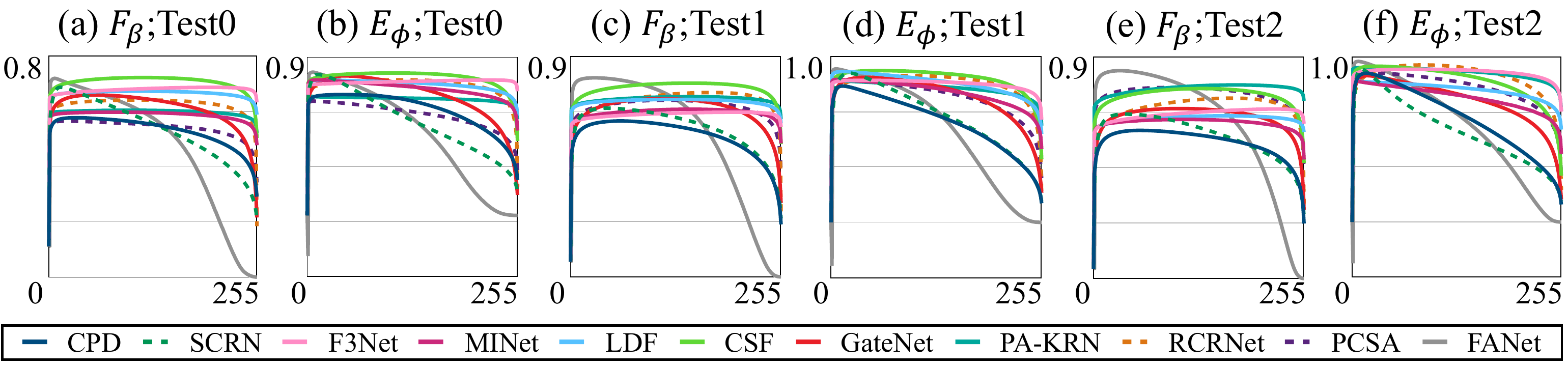}
    \end{overpic}
	\caption{F-measure ($F_{\beta}$) and E-measure ($E_\phi$) curves of all \OurTotalBaselines~baselines upon our \ourdataset.}
    \label{fig:curves_classes}
\end{figure*}

\begin{figure*}[t!]
	\centering
	\begin{overpic}[width=0.99\textwidth]{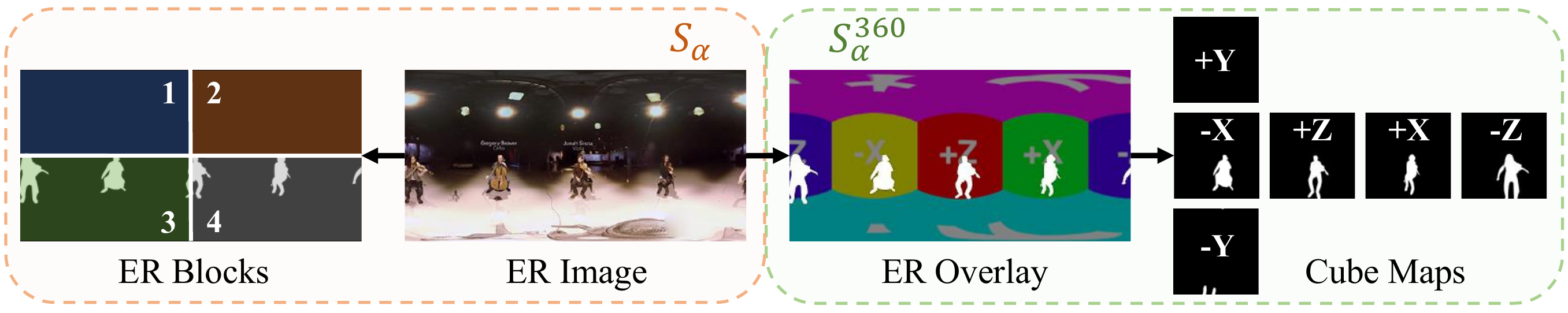}
    \end{overpic}
	\caption{A comparison between traditional S-measure ($S_\alpha$) \cite{Fan2017Smeasure} and proposed 360° geometry-adapted S-measure ($S_\alpha^{360}$). The $S_\alpha$/$S_\alpha^{360}$ compute region similarities ($S_r$/$S_r^{360}$) based on ER blocks/cube maps, respectively. '+X','-X','+Y','-Y','+Z' and '-Z' denote cube maps covering a FoV of 90°$\times$90°, observed from the right, left, up, down, front and back by a 360° camera.}
    \label{fig:metrics}
\end{figure*}

\section{Benchmark Experiments}\label{sec:benchmark}
\subsection{Experimental Settings}\label{sec:experiment_setting}
\noindent
\textbf{General Metrics}. In this work, we apply four widely used SOD metrics to the quantitative comparison of \OurTotalVideos~baselines, which are all representative SOTA SOD/VSOD models. Specifically, we adopt the recently proposed S-measure ($S_{\alpha}$) \cite{Fan2017Smeasure} and E-measure ($E_{\phi}$) \cite{Fan2018Enhanced}, commonly agreed Mean Absolute Error ($\mathcal{M}$) \cite{MAE} and F-measure ($F_{\beta}$) \cite{Fmeasure}. %The F-measure and MAE focus on the local (per-pixel) match between ground truth and prediction, while S-measure shifts attention to the evaluation of object structure and local region similarities between ground truth and saliency prediction maps. E-measure, a cognitive vision-inspired metric, considers both the local and global similarities by computing an enhanced alignment matrix ($\phi$). 
Please refer to appendices for detailed descriptions.
\par
\noindent
\textbf{360° Geometry-adapted S-measure}. Traditional $S_{\alpha}$ \eqnref{equ:sm} \cite{Fan2017Smeasure} proposed for SOD in 2D images with center-biased foreground objects.
\begin{equation}\label{equ:sm}
   S_\alpha = \alpha S_{o}(P,G) + (1 - \alpha) S_{r}(P,G),
\end{equation}
where $\alpha=0.5$, $S_{o}(\cdot)$ and $S_{r}(\cdot)$ denote the object/region similarities between each of the saliency prediction map ($P$) and the ground-truth map ($G$). To compute $S_{r}(\cdot)$, each $\{P,G\}$ pair is first divided into four blocks by using a horizontal and a vertical cut-off lines. However, natural ER images own a FoV of 360°$\times$180°, thus usually containing smaller salient objects distributed near the image equator. The significant divergence of objects' distribution and size may lead to inappropriate computation of $S_r(\cdot)$ based on ER blocks (Figure \ref{fig:metrics}). Considering the unique geometrical features of ER images, we further propose a 360° geometry-adapted S-measure ($S_\alpha^{360}$):
\begin{equation}\label{equ:sm_360}
   S_\alpha^{360} = \alpha S_{o}(P,G) + (1 - \alpha) S_{r}^{360}, \text{where}~S_{r}^{360} = \sum_{m=1}^{M}\omega_{m} ssim_{m}(P,G),
\end{equation}
$\omega_{m}=1/6$, $M$ denotes total number of cube maps, $S_{r}^{360}$ is computed based on $ssim_{m}(\cdot), m\in M$ \cite{wang2004image} on each of the cube maps ($\{+X,-X,+Y,-Y,+Z,-Z\}$ (Figure \ref{fig:metrics})). 
\par
\noindent
\textbf{Benchmark Models}. To contribute the community a comprehensive benchmark study, as well as filling the blank of 360°-VSHD, we collect 8/2/1 open-sourced SOTA SOD/VSOD/360°-SOD methods as the baselines. For SOD, we select well-known CPD \cite{CPD}, SCRN \cite{SCRN} and F3Net \cite{F3Net}, recently proposed MINet \cite{MINet}, LDF \cite{CVPR2020LDF}, CSF \cite{SOD100K}, GateNet \cite{GateNet} and PA-KRN\cite{xu2021locate}. For VSOD, we notice exclusively designed algorithms including RCRNet \cite{RCRNet} and PCSA \cite{gu2020PCSA}. We further adopt one newly proposed 360°-SOD method, FANet \cite{huang2020fanet}. Note that all the benchmark models are selected based on four prerequisites. The chosen methods must 1) own easily applied end-to-end structures. 2) be recent published SOTA methods evaluated on widely recognized SOD/VSOD benchmark datasets. 3) provide official training as well as inference codes with well written documents to ensure accurate re-implementation. 4) all based on ImageNet \cite{deng2009imagenet} pre-training.  
\par
\noindent
\textbf{Training Protocol}. To ensure fair comparison of \OurTotalBaselines~baselines, we re-train all models using only the training set of our \ourdataset, from their initial checkpoints pre-trained on ImageNet \cite{deng2009imagenet}. Note that we re-train all \OurTotalBaselines~benchmark models based on officially released codes with recommended hyperparameters. We conduct all experiments based on a platform which consists of Intel$^\circledR$ Xeon(R) W-2255 CPU @ 3.70GHz and one Quadro RTX 6000 GPU.

\subsection{Performance Comparison}\label{sec:comparison}
\noindent
\textbf{General Performance}. The quantitative results of all 11 baselines over
%three testing sets of
our \ourdataset~are shown in Table \ref{tab:QuantitativeComparison}. Generally, these SOD/VSOD methods show a gap between their performance on \ourdataset~and on existing SOD benchmarks (\eg, 
%GateNet \cite{GateNet} on \ourdataset-test0/DUTS-TE: $F_{\beta}=0.660<0.888$, CSF \cite{SOD100K} on \ourdataset-test1/DUTS-TE:$F_{\beta}=0.791<0.893$, 
RCRNet \cite{RCRNet} on \ourdataset-test2/DAVIS2016: $S_{\alpha}=0.828<0.884$, the mean $F_\beta$ scores of all baselines on \ourdataset~are under 0.8), which indicates the strong challenges brought by our \ourdataset~when compared to 2D datasets.
Besides, as for 360°-SOD method, FANet \cite{huang2020fanet} shows relatively high maximum $F_{\beta}$/$E_{\phi}$ scores across all testing sets (Figure \ref{fig:curves_classes}). However, a rapid decline of $F_{\beta}$ and $E_{\phi}$ scores appears as image intensity threshold increases. 
\par
\noindent
\textbf{Attributes-based Performance}. To help assess the model performance over different types of parameters and variations, we further evaluate all baselines based on each of the six attributes of our \ourdataset~(Figure \ref{fig:categories} (f)).
%As shown in Table \ref{tab:QuantitativeComparison} and Table \ref{tab:AttributesComparison},
%A consistency between the results of general and attributes-based comparison is observed according to Table \ref{tab:QuantitativeComparison} and {\color{cyan} SM} . In other words, F3Net \cite{F3Net}, LDF \cite{CVPR2020LDF}, CSF \cite{SOD100K}, RCRNet \cite{RCRNet}, PCSA \cite{gu2020PCSA} and FANet \cite{huang2020fanet} show relatively high performance when compared to other baselines.
Please refer to appendices for details. 

\begin{figure*}[t!]
	\centering
	\begin{overpic}[width=0.99\textwidth]{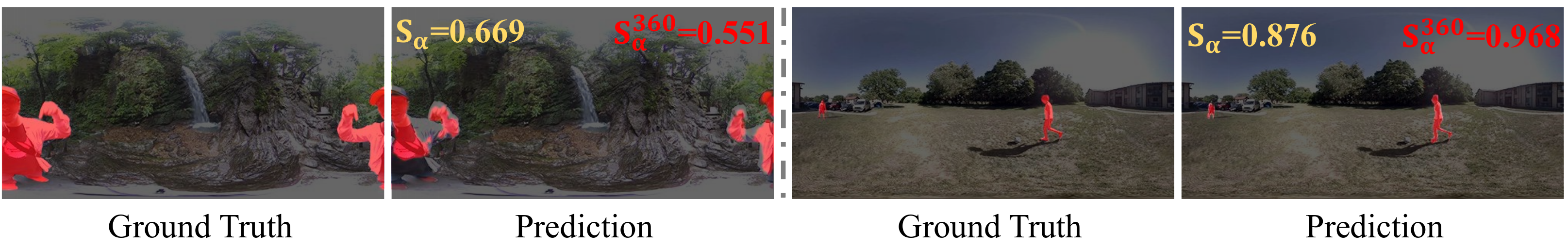}
    \end{overpic}
	\caption{Visual results evaluated with different metrics. $S_\alpha$ = S-measure \cite{Fan2017Smeasure}, $S_\alpha^{360}$ = proposed 360° Geometry-adapted S-measure.}
    \label{fig:metrics_analysis}
\end{figure*}

\section{Discussion}\label{sec:discussion}
In this section, we present some interesting findings about our dataset and benchmark.
\par
\noindent
\textbf{Domain Gap}.
As discussed in \secref{sec:comparison}, a gap between the performance of the SOTA SOD/VSOD methods on the existing datasets and our \ourdataset~is spotted. The finding indicates the significant challenges for 360°-SOD/-VSOD, which is consistent with recent 360°-SOD researches \cite{li2019distortion, Yi2020fSOD,ma2020stage}. To take a deeper look at 360°-VSHD, we design and attach different attributes \secref{sec:data_attributes} to each of the \OurTotalVideos~360° video scenes. These attributes highlight both the general (\emph{MP}, \emph{MB}, \emph{OC}, \emph{DV}) and 360°-specific (\emph{GD}, \emph{OV}) challenges for object segmentation tasks. By providing further quantitative comparison from the perspective of attributes (\secref{sec:comparison}), we aim to disentangle the 360°-VSHD to six sub-issues regarding omnidirectional image segmentation. Therefore, the key issues may include small object detection, ER projection-induce distortion mitigation, multi-projection-based object localization in 360°$\times$180° FoV, moving object detection, occlusion reasoning and multiple foreground objects ranking/segmentation. So far, our benchmark studies only focus on binary segmentation. 
%Future works may further explore the 360°-VSHD by considering above mentioned issues. 
\par
\begin{wrapfigure}{r}{0.5\textwidth}
\vspace{-6.5mm}
  \begin{center}
    \includegraphics[width=0.49\textwidth]{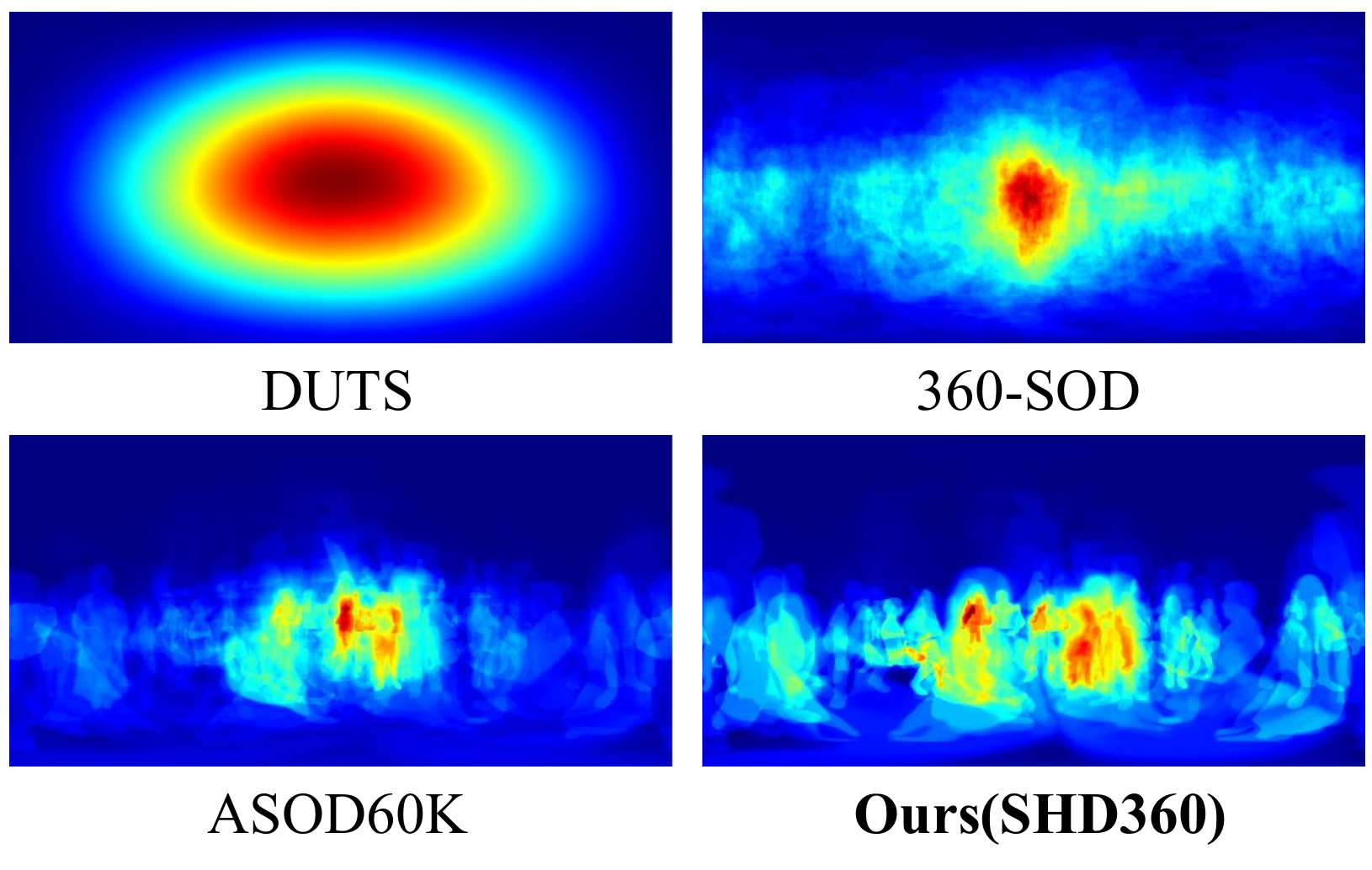}
  \end{center}
  \vspace{-14pt}
  \caption{Ground truth distribution of different datasets, including DUTS \cite{DUTS}, 360-SOD \cite{li2019distortion}, ASOD60K \cite{zhang2021asod60k} and our \ourdataset.}
 \label{fig:dataset_bias}
\end{wrapfigure}

\noindent
\textbf{Dataset Bias}. As shown in Figure \ref{fig:dataset_bias}, 360° datasets tend to show a equator bias, rather than only center bias existed in most of the 2D SOD datasets (\eg, DUTS \cite{DUTS}). Besides, our \ourdataset~shows relatively weaker center bias compared to other 360° datasets. The phenomenon is due to the photographers' tendency of framing the salient targets parallel to the 360° cameras. Modeling such a bias may help improve performance and aid epistemic uncertainty estimation for explainable deep learning.

\par
\noindent
\textbf{Evaluation Methods}. To the best of our knowledge, so far there is no exclusive metric for the quantitative assessment of 360°-SOD. In addition, the only structure-focused SOD metric, S-measure \cite{Fan2017Smeasure}, may not well adapted to the evaluation in ER images (Figure \ref{fig:metrics}). The ER image captures a FoV of 360°$\times$180° thus owning large image area as background scenes, while relatively small object regions randomly distributed near the equator. However, the S-measure is designed for the situation where one or a few obvious foreground objects with appropriate size distributed near the center of the given 2D image. As a result, our $S_{\alpha}^{360}$ (\eqnref{equ:sm_360}) is proved to be more sensitive to the evaluation on 360° images (Figure \ref{fig:metrics_analysis}). The $S_{\alpha}^{360}$ is able to give high score when the prediction is close to the ground truth, vice versa. The high sensitivity of our $S_{\alpha}^{360}$ is benefited from the cube map-based region similarity computation, which inspires future works toward fair comparison of object segmentation methods applied in 360° images and videos.

%more appropriate for evaluation on ER images, by computing region similarities based on cube maps which contain intact salient objects (Figure \ref{fig:metrics}). Future works may propose new metrics regarding the 360° geometry for fair comparison between 360°-SOD/-VSOD algorithms.
%\par
%\noindent
%\textbf{3D Benchmark}. As the development of RGB-D (or 3D) SOD benchmarks \cite{zhou2021rgb}, 3D SOD is becoming an emerging research field. Similarly, with the development of 3D VR cameras such as Insta360 Pro2, 3D VR SOD may be a promising as well as interesting topic to explore.

%\section{Conclusion}\label{sec:conclusion}
%In this paper, we establish \ourdataset, the first human-centric 360° VSOD dataset, which contains \OurTotalFrames~video frames reflecting \OurTotalVideos~real world daily scenes with attached attributes, and \OurTotalObjects/\OurTotalInstances~manually labeled object-/instance-level per-pixel annotations. Besides, we conduct extensive experiments involving \OurTotalBaselines~baselines and five metrics. Finally, based on our benchmark studies, we present a discussion and shed new light on future development of 360° SOD/VSOD.

\clearpage
\begin{appendices}

\begin{abstract}
In this document, we provide details about the proposed \ourdataset~and the new benchmark.
\begin{itemize}  
  \item \textbf{Evaluation Metrics.} In \secref{sec:metrics}, we describe the details of three widely applied evaluation metrics including F-measure \cite{Fmeasure}, enhanced-alignment measure (E-measure) \cite{Fan2018Enhanced} and mean absolute error (MAE) \cite{MAE}. Please refer to our manuscript for details of the structure measure (S-measure) \cite{Fan2017Smeasure} and the newly proposed 360° geometry-adapted S-measure.
  
  %\item \textbf{Dataset.} In \secref{sec:dataset}, we introduce detailed information about the proposed \ourdataset, from the aspects of dataset usage, annotations and attributes.
  
  \item \textbf{Experiments.} In \secref{sec:experiment}, we show extensive quantitative and qualitative results of \OurTotalBaselines~baselines upon our \ourdataset.
\end{itemize}
%Please note that our dataset and benchmark will be released at \url{https://github.com/PanoAsh/SHD360}.
\end{abstract}

\section{General Metrics}\label{sec:metrics}
In this work, we evaluate all \OurTotalBaselines~benchmark models with four widely used salient object detection (SOD) metrics and our proposed 360° geometry-adapted S-measure ($S_\alpha^{360}$), with respect to the ground-truth binary map and predicted saliency map. The F-measure ($F_\beta$) \cite{Fmeasure} and mean absolute error (MAE) \cite{MAE} focus on the local (per-pixel) match between ground truth and prediction, while S-measure ($S_\alpha$) \cite{Fan2017Smeasure} and $S_\alpha^{360}$ pay attention to the object structure similarities (please refer to the manuscript for details). Besides, E-measure ($E_\phi$) \cite{Fan2018Enhanced} considers both the local and global information.
\par
\noindent
$\bullet$ \textbf{MAE} computes the mean absolute error between the ground truth $G \in \{0, 1\}$ and a normalized predicted saliency map $P \in [0, 1]$, i.e.,
\begin{equation}\label{equ:mae}
   MAE = \frac{1}{W\times{H}}\sum_{i=1}^{W}\sum_{j=1}^{H}\mid G(i, j) - P(i, j)\mid,
\end{equation}
where $H$ and $W$ denote height and width, respectively.
\par
\noindent
$\bullet$ \textbf{F-measure} gives a single value ($F_{\beta}$) considering both the $Precision$ and $Recall$, which is defined as:
\begin{equation}\label{equ:fm}
   F_{\beta} = \frac{(1+\beta^{2})Precision \times Recall}{\beta^{2}Precision + Recall}, 
\end{equation}
with
\begin{equation}
  Precision=\frac{\left|P\cap G\right|}{\left|P\right|}; Recall=\frac{\left|P\cap G\right|}{\left|G\right|},
\end{equation}
where $P$ denotes a binary mask converted from a predicted saliency map and $G$ is the ground truth. Multiple $P$ are computed by assigning different thresholds $\tau, \tau \in [0,255]$ on the predicted saliency map. Note that the $\beta^{\text{2}}$ is set to 0.3 according to \cite{Fmeasure}.
\par
\noindent
$\bullet$ \textbf{E-measure} is a cognitive vision-inspired metric to evaluate both the local and global similarities between two binary maps. Specifically, it is defined as:
\begin{equation}\label{equ:em}
E_{\phi}=\frac{1}{W\times H}\sum_{x=1}^W\sum_{y=1}^H\phi\left(P(x,y), G(x,y)\right),
\end{equation} 
where $\phi$ represents the enhanced alignment matrix \cite{Fan2018Enhanced}.

%\section{Dataset}\label{sec:dataset}
%The raw videos of our \ourdataset~were downloaded from YouTube with permissions from corresponding owners. Please note that our \ourdataset~is released for academic use only, with a license of creative commons (CC BY-NC-SA 3.0, \url{https://creativecommons.org/licenses/by-nc-sa/3.0/}). As stated in the manuscript $\S 3.3$, we conducted strict quality control to ensure high-quality annotations. Examples are shown in Figure \ref{fig:pass_reject}. Besides, we present visual examples of all sequences of our \ourdataset~in Figure \ref{fig:annotation_sum}. Please refer to \url{https://github.com/PanoAsh/SHD360} for the complete annotations. Finally, we list detailed sequence-attached attributes in Table \ref{tab:attributes_details}.

\section{Experiments}\label{sec:experiment}
\begin{figure*}[t!]
	\centering
	\begin{overpic}[width=0.99\textwidth]{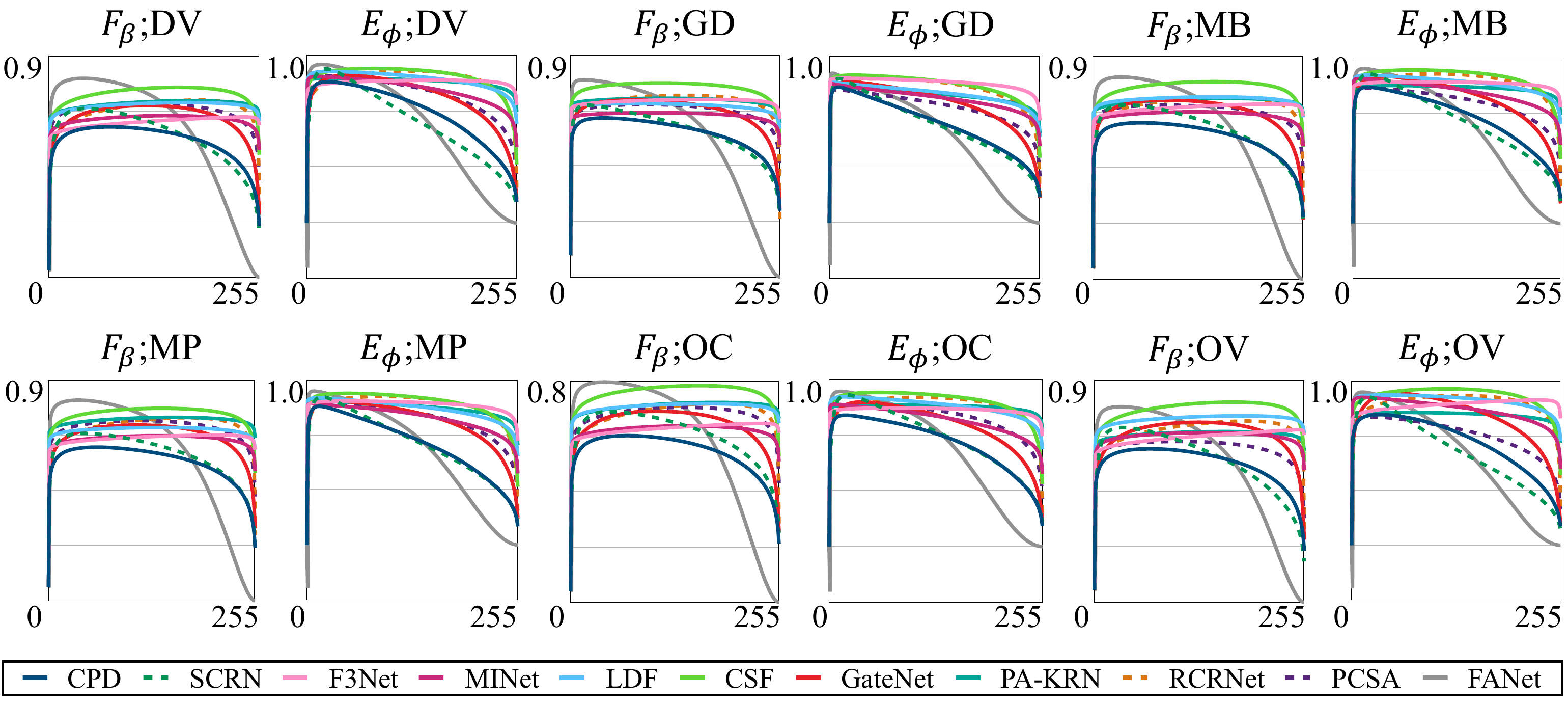}
    \end{overpic}
	\caption{Attributes-specific F-measure ($F_{\beta}$) and E-measure ($E_\phi$) curves of all \OurTotalBaselines~baselines upon our \ourdataset.}
    \label{fig:curves_attributes}
\end{figure*}

\begin{figure*}[t!]
	\centering
	\begin{overpic}[width=0.99\textwidth]{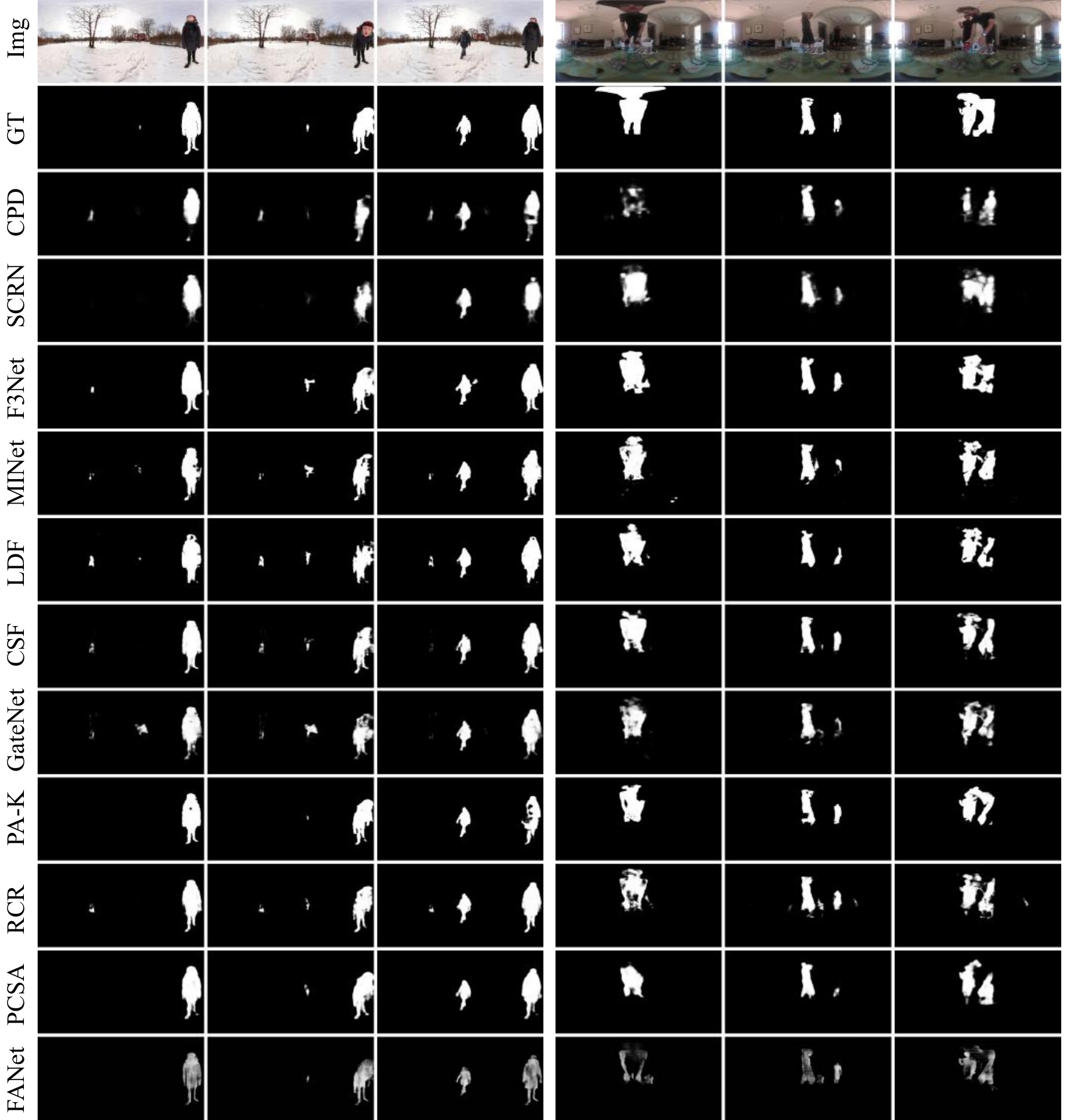}
    \end{overpic}
	\caption{Visual results of all baselines upon \ourdataset-test0. Img = image. GT = ground truth.}
    \label{fig:visual_te0}
\end{figure*}

\begin{figure*}[t!]
	\centering
	\begin{overpic}[width=0.99\textwidth]{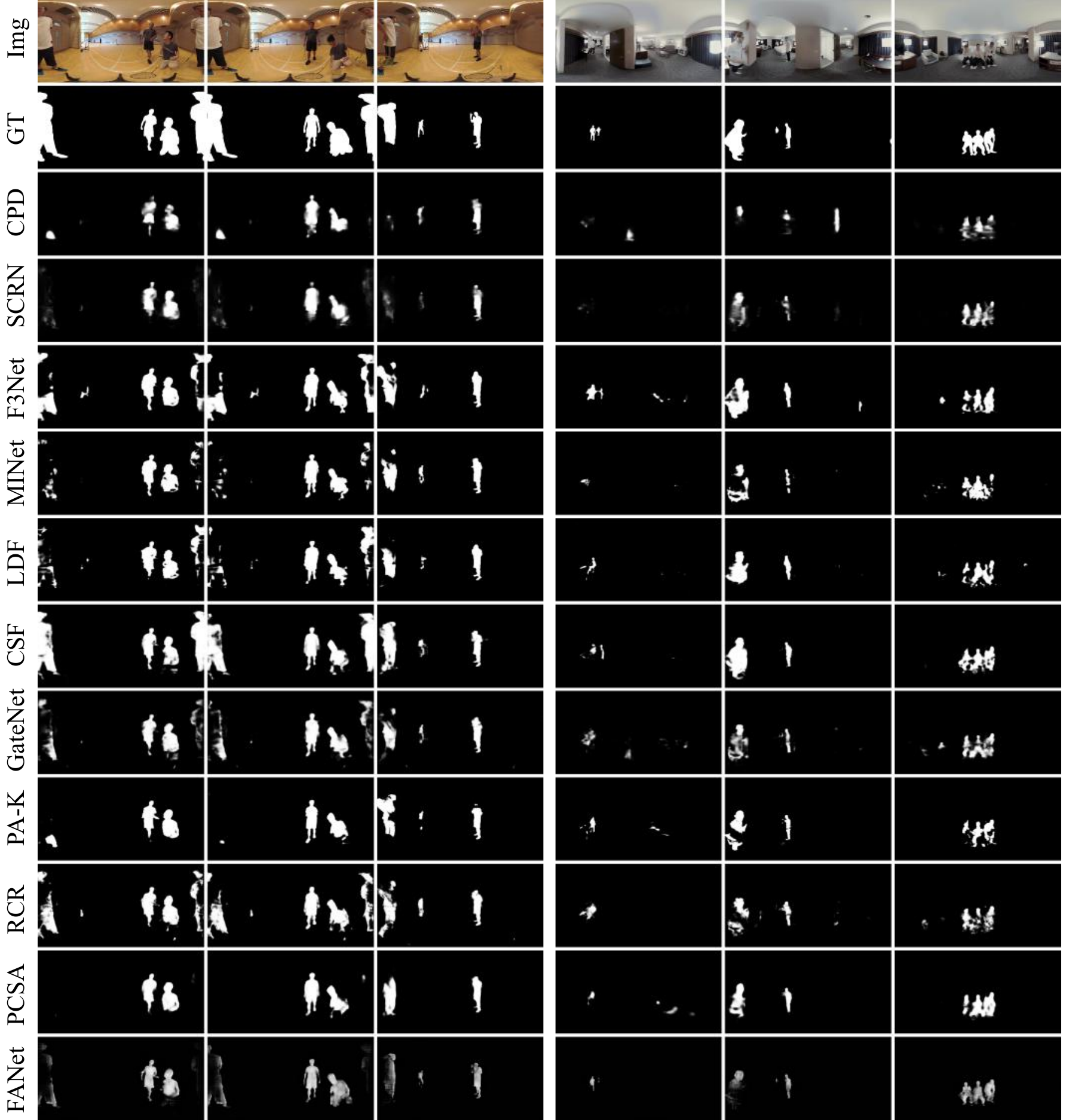}
    \end{overpic}
	\caption{Visual results of all baselines upon \ourdataset-test1. Img = image. GT = ground truth.}
    \label{fig:visual_te1}
\end{figure*}

\begin{figure*}[t!]
	\centering
	\begin{overpic}[width=0.99\textwidth]{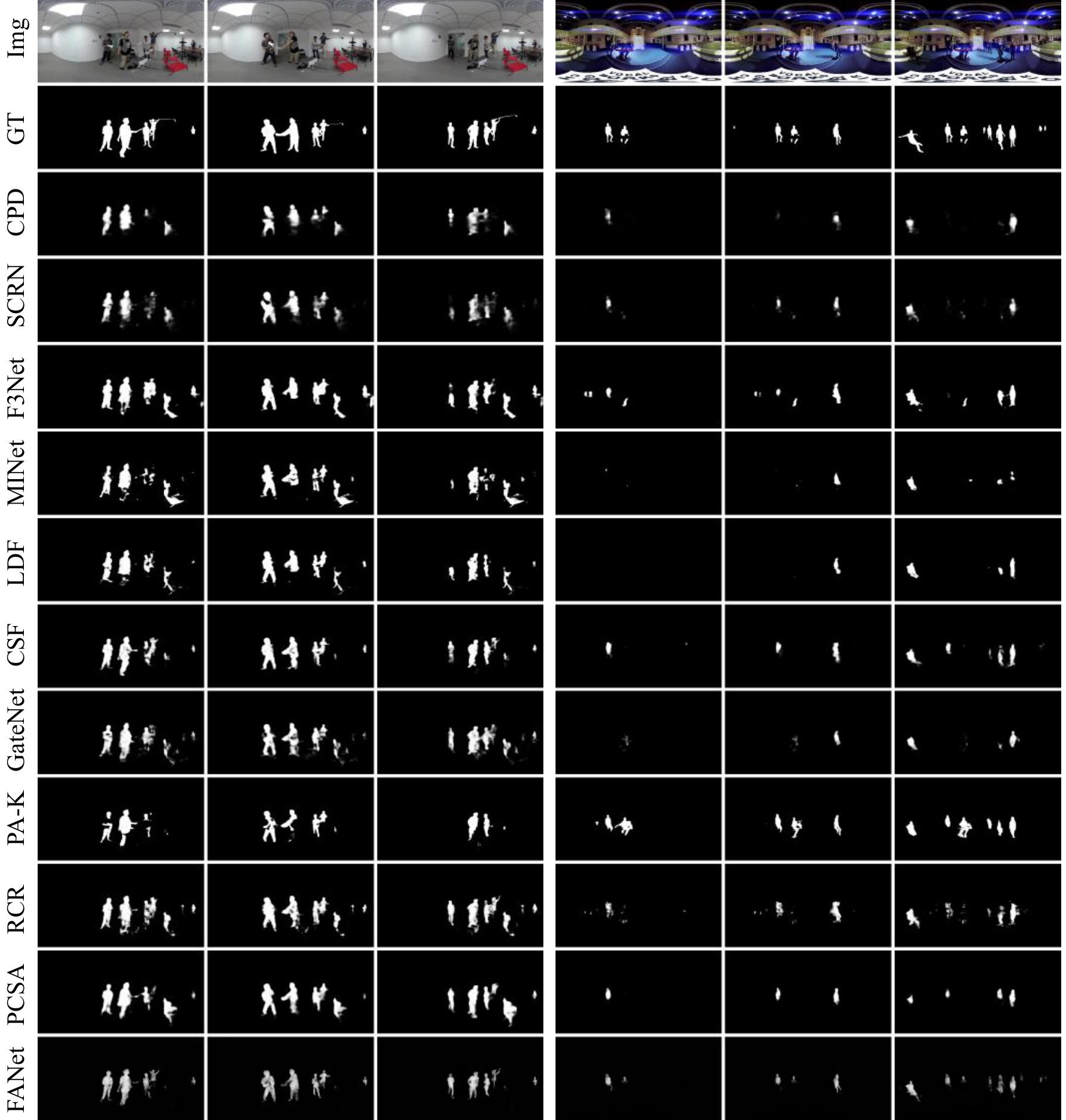}
    \end{overpic}
	\caption{Visual results of all baselines upon \ourdataset-test2. Img = image. GT = ground truth.}
    \label{fig:visual_te2}
\end{figure*}

In this section. we provide extensive results of our benchmark studies. Figure \ref{fig:curves_attributes} shows attributes-specific F-/E-measure curves regarding all 11 baselines. Table \ref{tab:AttributesComparison} presents attributes-dependent quantitative result, which highlights relative good performance of baselines such as F3Net \cite{F3Net}, LDF \cite{CVPR2020LDF}, CSF \cite{SOD100K}, PA-KRN \cite{xu2021locate}, RCRNet \cite{RCRNet} and PCSA \cite{gu2020PCSA}. Besides, we also provide sequence-based quantitative results in Table \ref{tab:seq_results}. Finally, we provide visual examples (Figure \ref{fig:visual_te0}, \ref{fig:visual_te1} and \ref{fig:visual_te2}) of qualitative results of all baselines upon our \ourdataset. Please also refer to `Demo-Outdoor-Tennis.mp4' for complete visual results of a sequence example, \ie, OutDoor-Tennis.

\begin{table*}[t!]
  \centering
  \renewcommand{\arraystretch}{1}
  \setlength\tabcolsep{3.5pt}
  \resizebox{0.99\textwidth}{!}{
  \begin{tabular}{rr||cccccccc||cc||c}
   \toprule
   &\multirow{3}{*}{Metrics} & \multicolumn{8}{c||}{SOD} & \multicolumn{2}{c||}{VSOD} & 360° SOD
  \\
  \cline{3-13}
   && ~~CPD~~  & ~SCRN~~  & ~F3Net~  & ~MINet~ & ~~LDF~~ & ~~CSF~~  & GateNet  & PA-KRN~   & RCRNet~  & ~PCSA~~  & ~FANet~
   \\
    && \cite{CPD} &  \cite{SCRN} &  \cite{F3Net} &  \cite{MINet} &  \cite{CVPR2020LDF} &  \cite{SOD100K} &  \cite{GateNet} &  \cite{xu2021locate}  &  \cite{RCRNet} &  \cite{gu2020PCSA} &  \cite{huang2020fanet}
    \\
  \midrule
  \multirow{5}{*}{MP} & $F_{\beta}~\uparrow$ & 
  %0.627 & 0.683 & 0.677 & 0.674 & 0.708 & {\color{blue} 0.786} & 0.717 & 0.540 & {\color{green} 0.748} & 0.736 & {\color{red} 0.819} \\
  0.568 & 0.601 & 0.662 & 0.659 & 0.696 & {\color{red} 0.761} & 0.672 & {\color{blue} 0.735} & {\color{green} 0.712} & 0.706 & 0.610 \\
                      & $S_{\alpha}~\uparrow$ & 0.695 & 0.713 & 0.773 & 0.741 & 0.768 & {\color{red} 0.835} & 0.775 & {\color{green} 0.797} & {\color{blue} 0.823} & 0.779 & 0.757 \\
                      & $S_{\alpha}^{360}~\uparrow$ & 0.719 & 0.648 & {\color{blue} 0.799} & 0.715 & 0.666 & 0.780 & 0.711 & {\color{red} 0.815}  & {\color{green} 0.792} & 0.788 & 0.642 \\
                      & $E_{\phi}~\uparrow$ & 
                      %0.882 & 0.925 & 0.905 & 0.894 & 0.920 & {\color{blue} 0.941} & 0.905 & 0.865 & {\color{green} 0.927} & 0.901 & {\color{red} 0.951} \\
                      0.710 & 0.714 & {\color{blue} 0.889} & 0.830 & 0.869 & {\color{red} 0.896} & 0.814 & 0.874 & {\color{green} 0.882} & 0.826 & 0.679 \\
                      & $\mathcal{M}~\downarrow$ & 0.031 & 0.029 & 0.024 & 0.026 & 0.025 & {\color{red} 0.017} & {\color{green} 0.023} & 0.024 & {\color{blue} 0.022} & 0.026 & 0.032 \\ 
  \midrule
  \multirow{5}{*}{OC} & $F_{\beta}~\uparrow$ & 
  %0.604 & 0.692 & 0.649 & 0.640 & {\color{green} 0.720} & {\color{blue} 0.786} & 0.693 & 0.524 & {\color{green} 0.720} & 0.707 & {\color{red} 0.799} \\
  0.546 & 0.608 & 0.628 & 0.624 & {\color{green} 0.706} & {\color{red} 0.758} & 0.642 & {\color{blue} 0.709} & 0.683 & 0.679 & 0.593 \\
                      & $S_{\alpha}~\uparrow$ & 0.691 & 0.718 & 0.754 & 0.719 & 0.767 & {\color{red} 0.829} & 0.755 & {\color{green} 0.794} & {\color{blue} 0.804} & 0.760 & 0.748 \\
                      & $S_{\alpha}^{360}~\uparrow$ & 0.708 & 0.632 & {\color{blue} 0.779} & 0.677 & 0.646 & {\color{green} 0.771} & 0.677 & {\color{red} 0.820} & 0.755 & 0.768 & 0.622 \\
                      & $E_{\phi}~\uparrow$ &
                      %0.840 & {\color{green} 0.931} & 0.872 & 0.889 & 0.922 & {\color{blue} 0.942} & 0.899 & 0.840 & 0.920 & 0.888 & {\color{red} 0.946} \\
                      0.716 & 0.741 & 0.861 & 0.818 & {\color{green} 0.878} & {\color{red} 0.902} & 0.796 & 0.876 & {\color{blue} 0.881} & 0.828 & 0.684 \\
                      & $\mathcal{M}~\downarrow$ & 0.024 & 0.022 & {\color{green} 0.018} & 0.020 & {\color{blue} 0.017} & {\color{red} 0.012} & 0.019 & {\color{green} 0.018} & {\color{blue} 0.017} & 0.020 & 0.025 \\ 
  \midrule
  \multirow{5}{*}{DV} & $F_{\beta}~\uparrow$ &
  %0.612 & 0.687 & 0.652 & 0.658 & 0.711 & {\color{blue} 0.773} & 0.699 & 0.536 & {\color{green} 0.721} & 0.707 & {\color{red} 0.809} \\
  0.555 & 0.577 & 0.628 & 0.640 & {\color{green} 0.696} & {\color{red} 0.742} & 0.648 & {\color{blue} 0.702} & 0.680 & 0.677 & 0.599 \\
                      & $S_{\alpha}~\uparrow$ & 0.714 & 0.714 & 0.775 & 0.751 & 0.782 & {\color{red} 0.835} & 0.778 & {\color{green} 0.798} & {\color{blue} 0.821} & 0.775 & 0.763 \\
                      & $S_{\alpha}^{360}~\uparrow$ & 0.729 & 0.634 & {\color{blue} 0.797} & 0.715 & 0.660 & 0.772 & 0.690 & {\color{red} 0.820} & 0.778 & {\color{green} 0.786} & 0.626  \\
                      & $E_{\phi}~\uparrow$ & 
                      %0.883 & {\color{green} 0.941} & 0.890 & 0.907 & 0.929 & {\color{blue} 0.942} & 0.912 & 0.870 & 0.932 & 0.908 & {\color{red} 0.960} \\
                      0.745 & 0.706 & 0.879 & 0.848 & {\color{blue} 0.893} & {\color{red} 0.903} & 0.816 & 0.882 & {\color{green} 0.889} & 0.828 & 0.689 \\
                      & $\mathcal{M}~\downarrow$ & 0.015 & 0.015 & 0.013 & 0.013 & {\color{green} 0.012} & {\color{red} 0.009} & 0.013 & {\color{green} 0.012} & {\color{blue} 0.011} & {\color{green} 0.012} & 0.017 \\ 
  \midrule
  \multirow{5}{*}{MB} & $F_{\beta}~\uparrow$ &
  %0.631 & 0.703 & 0.707 & 0.675 & {\color{green} 0.735} & {\color{blue} 0.797} & 0.723 & 0.530 & 0.733 & 0.702 & {\color{red} 0.815} \\
  0.573 & 0.606 & 0.690 & 0.660 & {\color{blue} 0.723} & {\color{red} 0.770} & 0.671 & {\color{green} 0.717} & 0.702 & 0.676 & 0.600 \\
                      & $S_{\alpha}~\uparrow$ & 0.707 & 0.719 & {\color{green} 0.793} & 0.747 & 0.781 & {\color{red} 0.846} & 0.773 & 0.790 & {\color{blue} 0.812} & 0.756 & 0.753 \\
                      & $S_{\alpha}^{360}~\uparrow$ & 0.714 & 0.624 & {\color{blue} 0.818} & 0.720 & 0.679 & 0.776 & 0.673 & {\color{red} 0.829} & 0.768 & {\color{green} 0.782} & 0.624 \\
                      & $E_{\phi}~\uparrow$ &
                      %0.867 & 0.927 & 0.891 & 0.906 & 0.922 & {\color{blue} 0.945} & 0.917 & 0.831 & {\color{green} 0.928} & 0.863 & {\color{red} 0.951} \\
                      0.729 & 0.716 & 0.882 & 0.843 & {\color{green} 0.883} & {\color{red} 0.909} & 0.813 & 0.861 & {\color{blue} 0.887} & 0.794 & 0.682 \\
                      & $\mathcal{M}~\downarrow$ & 0.025 & 0.024 & {\color{blue} 0.016} & 0.020 & 0.018 & {\color{red} 0.012} & 0.019 & 0.020 & {\color{green} 0.017} & 0.021 & 0.027 \\
  \midrule
  \multirow{5}{*}{OV} & $F_{\beta}~\uparrow$ &
  %0.623 & 0.709 & 0.701 & 0.688 & {\color{green} 0.757} & {\color{red} 0.812} & 0.731 & 0.533 & 0.735 & 0.653 & {\color{blue} 0.794} \\
  0.562 & 0.579 & 0.668 & 0.672 & {\color{blue} 0.741} & {\color{red} 0.781} & 0.680 & 0.681 & {\color{green} 0.697} & 0.625 & 0.577 \\
                      & $S_{\alpha}~\uparrow$ & 0.702 & 0.710 & 0.792 & 0.755 & {\color{green} 0.806} & {\color{red} 0.865} & 0.791 & 0.760 & {\color{blue} 0.816} & 0.720 & 0.734 \\
                      & $S_{\alpha}^{360}~\uparrow$ & 0.698 & 0.628 & {\color{green} 0.791} & 0.701 & 0.687 & {\color{red} 0.803} & 0.696 & {\color{blue} 0.801} & 0.766 & 0.749 & 0.623 \\
                      & $E_{\phi}~\uparrow$ & 
                      %0.849 & 0.930 & 0.916 & 0.928 & 0.939 & {\color{red} 0.967} & {\color{green} 0.946} & 0.821 & 0.936 & 0.838 & {\color{blue} 0.952} \\
                      0.722 & 0.682 & 0.892 & 0.869 & {\color{blue} 0.911} & {\color{red} 0.931} & 0.840 & 0.846 & {\color{green} 0.896} & 0.757 & 0.664 \\
                      & $\mathcal{M}~\downarrow$ & 0.029 & 0.028 & {\color{blue} 0.020} & 0.023 & {\color{blue} 0.020} & {\color{red} 0.013} & 0.022 & 0.025 & {\color{green} 0.021} & 0.027 & 0.032 \\ 
  \midrule
  \multirow{5}{*}{GD} & $F_{\beta}~\uparrow$ & 
  %0.645 & 0.698 & 0.721 & 0.667 & 0.704 & {\color{blue} 0.788} & 0.718 & 0.536 & {\color{green} 0.737} & 0.700 & {\color{red} 0.800} \\
  0.585 & 0.596 & {\color{green} 0.711} & 0.655 & 0.693 & {\color{red} 0.765} & 0.672 & {\color{blue} 0.715} & {\color{green} 0.711} & 0.674 & 0.579 \\
                      & $S_{\alpha}~\uparrow$ & 0.696 & 0.703 & {\color{green} 0.786} & 0.735 & 0.754 & {\color{red} 0.824} & 0.765 & 0.762 & {\color{blue} 0.801} & 0.749 & 0.736 \\
                      & $S_{\alpha}^{360}~\uparrow$ & 0.701 & 0.620 & {\color{red} 0.810} & 0.721 & 0.666 & 0.765 & 0.675 & {\color{blue} 0.798} & 0.757 & {\color{green} 0.770} & 0.619 \\
                      &  $E_{\phi}~\uparrow$ &
                      %0.860 & 0.898 & {\color{green} 0.900} & 0.875 & 0.884 & {\color{blue} 0.914} & 0.883 & 0.797 & 0.898 & 0.848 & {\color{red} 0.922} \\
                      0.694 & 0.689 & {\color{red} 0.878} & 0.807 & 0.828 & {\color{blue} 0.866} & 0.785 & 0.822 & {\color{green} 0.850} & 0.770 & 0.643 \\
                      & $\mathcal{M}~\downarrow$ & 0.055 & 0.054 & {\color{blue} 0.043} & 0.050 & 0.048 & {\color{red} 0.040} & 0.048 & 0.050 & {\color{green} 0.046} & 0.051 & 0.058 \\ 
  \bottomrule
  \end{tabular}}
   \caption{Attributes-based performance comparison of \OurTotalBaselines~baselines over our \ourdataset. $S_\alpha$ = S-measure ($\alpha$=0.5 \cite{Fan2017Smeasure}), $S_{\alpha}^{360}$ = 360° geometry-adapted S-measure, $F_\beta$ = 
  % maximum F-measure
  mean F-measure
   ($\beta^2$=0.3) \cite{Fmeasure}, $E_\phi$ = 
  % maximum E-measure
  mean E-measure
   \cite{Fan2018Enhanced}, $\mathcal{M}$ = mean absolute error \cite{MAE}. $\uparrow$/$\downarrow$ denotes a larger/smaller value is better. Three best results of each row are in {\color{red} \textbf{red}}, {\color{blue} \textbf{blue}} and {\color{green} \textbf{green}}, respectively.}
   \label{tab:AttributesComparison}
\end{table*}

\begin{figure*}[t!]
	\centering
	\begin{overpic}[width=0.99\textwidth]{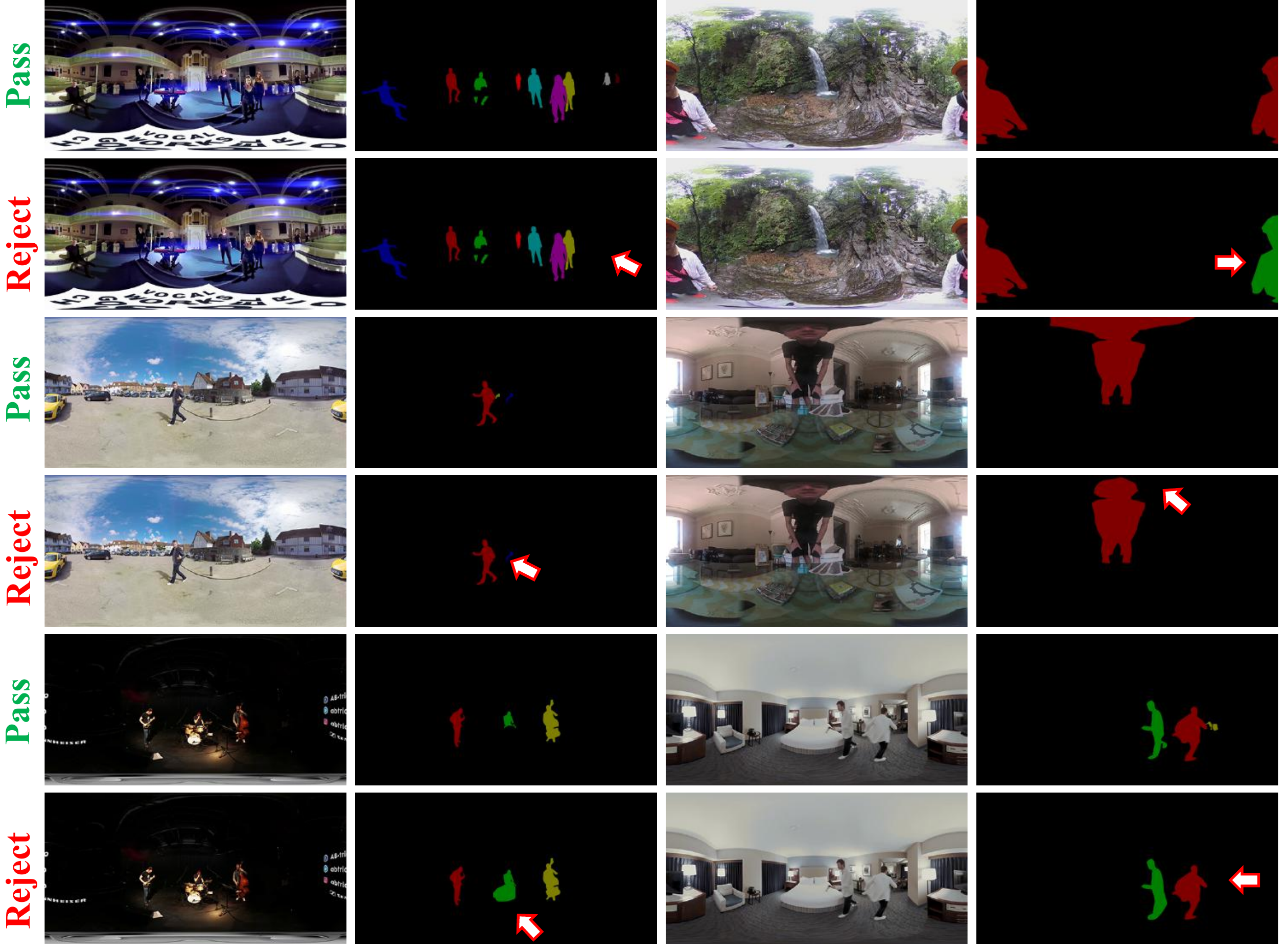}
    \end{overpic}
	\caption{Passed and rejected examples of annotation quality control process.}
    \label{fig:pass_reject}
\end{figure*}

\begin{table*}[t!]
  \centering
  \renewcommand{\arraystretch}{1}
  \setlength\tabcolsep{1.2pt}
  \resizebox{0.99\textwidth}{!}{
  \begin{tabular}{l|r||cccccccc|cc|c}
   \toprule
   Super-class/ &\multirow{2}{*}{Metrics} & \multicolumn{8}{c|}{SOD} & \multicolumn{2}{c|}{VSOD} & \multicolumn{1}{c}{360° SOD}
  \\
  \cline{3-13}
   Sub-class && CPD \cite{CPD}~ & SCRN \cite{SCRN}~ & F3Net \cite{F3Net}~ & MINet \cite{MINet}~ & LDF \cite{CVPR2020LDF}~ & CSF \cite{SOD100K}~ & GateNet \cite{GateNet}~ & PA-KRN \cite{xu2021locate}~ & RCRNet \cite{RCRNet}~ & PCSA \cite{gu2020PCSA}~ & FANet \cite{huang2020fanet}
   \\
  \hline
  \multirow{4}{*}{O./Japanese} 
    & $S_{\alpha}~\uparrow$ & 0.706 & 0.766 & 0.847 & 0.775 & 0.834 & 0.924 & 0.823 & 0.625 & 0.744 & 0.535 & 0.654 \\
    & max $E_{\phi}~\uparrow$ & 0.952 & 0.982 & 0.981 & 0.973 & 0.967 & 0.989 & 0.983 & 0.799 & 0.973 & 0.624 & 0.987 \\
    & mean $E_{\phi}~\uparrow$ & 0.637 & 0.664 & 0.947 & 0.934 & 0.904 & 0.951 & 0.865 & 0.685 & 0.880 & 0.464 & 0.559 \\
    & $\mathcal{M}~\downarrow$ & 0.047 & 0.044 & 0.019 & 0.033 & 0.025 & 0.017 & 0.030 & 0.043 & 0.039 & 0.056 & 0.054 \\ 
  \hline
  \multirow{4}{*}{I./SingingDancing} 
    & $S_{\alpha}~\uparrow$ & 0.594 & 0.620 & 0.671 & 0.562 & 0.601 & 0.700 & 0.601 & 0.842 & 0.725 & 0.644 & 0.712 \\
    & max $E_{\phi}~\uparrow$ & 0.967 & 0.982 & 0.981 & 0.724 & 0.683 & 0.983 & 0.983 & 0.968 & 0.986 & 0.909 & 0.991 \\
    & mean $E_{\phi}~\uparrow$ & 0.511 & 0.550 & 0.807 & 0.432 & 0.493 & 0.776 & 0.503 & 0.950 & 0.822 & 0.620 & 0.623 \\
    & $\mathcal{M}~\downarrow$ & 0.015 & 0.015 & 0.014 & 0.015 & 0.014 & 0.012 & 0.015 & 0.010 & 0.013 & 0.013 & 0.016 \\ 
    \hline
  \multirow{4}{*}{O./Walking} 
    & $S_{\alpha}~\uparrow$ & 0.923 & 0.757 & 0.959 & 0.952 & 0.812 & 0.968 & 0.886 & 0.759 & 0.899 & 0.891 & 0.816 \\
    & max $E_{\phi}~\uparrow$ & 0.995 & 0.989 & 0.994 & 0.994 & 0.970 & 0.996 & 0.992 & 0.869 & 0.993 & 0.985 & 0.995 \\
    & mean $E_{\phi}~\uparrow$ & 0.939 & 0.674 & 0.985 & 0.976 & 0.859 & 0.980 & 0.854 & 0.742 & 0.916 & 0.894 & 0.706 \\
    & $\mathcal{M}~\downarrow$ & 0.011 & 0.026 & 0.006 & 0.008 & 0.019 & 0.006 & 0.018 & 0.025 & 0.014 & 0.016 & 0.027 \\ 
    \hline
  \multirow{4}{*}{I./Melodrama} 
    & $S_{\alpha}~\uparrow$ & 0.744 & 0.774 & 0.823 & 0.793 & 0.764 & 0.861 & 0.858 & 0.773 & 0.830 & 0.792 & 0.747 \\
    & max $E_{\phi}~\uparrow$ & 0.896 & 0.951 & 0.900 & 0.889 & 0.860 & 0.958 & 0.966 & 0.852 & 0.943 & 0.863 & 0.947 \\
    & mean $E_{\phi}~\uparrow$ & 0.678 & 0.729 & 0.825 & 0.781 & 0.734 & 0.856 & 0.857 & 0.751 & 0.816 & 0.755 & 0.644 \\
    & $\mathcal{M}~\downarrow$ & 0.113 & 0.102 & 0.083 & 0.096 & 0.105 & 0.071 & 0.071 & 0.103 & 0.082 & 0.098 & 0.122 \\ 
    \hline
  \multirow{4}{*}{I./Studio} 
    & $S_{\alpha}~\uparrow$ & 0.690 & 0.675 & 0.839 & 0.783 & 0.808 & 0.832 & 0.797 & 0.814 & 0.868 & 0.888 & 0.760 \\
    & max $E_{\phi}~\uparrow$ & 0.977 & 0.983 & 0.984 & 0.986 & 0.988 & 0.980 & 0.988 & 0.984 & 0.988 & 0.991 & 0.993 \\
    & mean $E_{\phi}~\uparrow$ & 0.764 & 0.655 & 0.969 & 0.892 & 0.947 & 0.907 & 0.871 & 0.925 & 0.934 & 0.925 & 0.667 \\
    & $\mathcal{M}~\downarrow$ & 0.018 & 0.017 & 0.012 & 0.013 & 0.013 & 0.012 & 0.014 & 0.011 & 0.012 & 0.009 & 0.018 \\ 
    \hline
  \multirow{4}{*}{O./RacingCar} 
    & $S_{\alpha}~\uparrow$ & 0.381 & 0.387 & 0.508 & 0.372 & 0.393 & 0.430 & 0.401 & 0.399 & 0.415 & 0.369 & 0.377 \\
    & max $E_{\phi}~\uparrow$ & 0.457 & 0.798 & 0.555 & 0.336 & 0.371 & 0.510 & 0.640 & 0.382 & 0.545 & 0.313 & 0.533 \\
    & mean $E_{\phi}~\uparrow$ & 0.281 & 0.297 & 0.441 & 0.281 & 0.304 & 0.346 & 0.311 & 0.330 & 0.349 & 0.274 & 0.278 \\
    & $\mathcal{M}~\downarrow$ & 0.265 & 0.262 & 0.236 & 0.267 & 0.259 & 0.249 & 0.258 & 0.257 & 0.258 & 0.269 & 0.269 \\ 
    \hline
  \multirow{4}{*}{I./Violins} 
    & $S_{\alpha}~\uparrow$ & 0.741 & 0.724 & 0.826 & 0.854 & 0.825 & 0.890 & 0.881 & 0.751 & 0.899 & 0.755 & 0.781 \\
    & max $E_{\phi}~\uparrow$ & 0.978 & 0.982 & 0.980 & 0.980 & 0.982 & 0.987 & 0.985 & 0.976 & 0.988 & 0.972 & 0.989 \\
    & mean $E_{\phi}~\uparrow$ & 0.761 & 0.659 & 0.960 & 0.951 & 0.927 & 0.946 & 0.907 & 0.846 & 0.936 & 0.792 & 0.662 \\
    & $\mathcal{M}~\downarrow$ & 0.027 & 0.028 & 0.023 & 0.020 & 0.022 & 0.017 & 0.019 & 0.023 & 0.016 & 0.025 & 0.027 \\ 
    \hline
  \multirow{4}{*}{O./Football} 
    & $S_{\alpha}~\uparrow$ & 0.680 & 0.652 & 0.763 & 0.720 & 0.809 & 0.807 & 0.757 & 0.825 & 0.763 & 0.593 & 0.687 \\
    & max $E_{\phi}~\uparrow$ & 0.890 & 0.926 & 0.945 & 0.859 & 0.943 & 0.946 & 0.918 & 0.938 & 0.929 & 0.798 & 0.936 \\
    & mean $E_{\phi}~\uparrow$ & 0.700 & 0.542 & 0.828 & 0.830 & 0.897 & 0.866 & 0.818 & 0.905 & 0.826 & 0.515 & 0.626 \\
    & $\mathcal{M}~\downarrow$ & 0.004 & 0.003 & 0.004 & 0.003 & 0.003 & 0.003 & 0.004 & 0.002 & 0.004 & 0.005 & 0.006 \\ 
    \hline
  \multirow{4}{*}{I./Director} 
    & $S_{\alpha}~\uparrow$ & 0.732 & 0.782 & 0.825 & 0.779 & 0.799 & 0.889 & 0.852 & 0.814 & 0.863 & 0.840 & 0.823 \\
    & max $E_{\phi}~\uparrow$ & 0.970 & 0.972 & 0.973 & 0.969 & 0.973 & 0.987 & 0.977 & 0.970 & 0.977 & 0.975 & 0.991 \\
    & mean $E_{\phi}~\uparrow$ & 0.799 & 0.826 & 0.956 & 0.945 & 0.920 & 0.935 & 0.898 & 0.885 & 0.940 & 0.944 & 0.755 \\
    & $\mathcal{M}~\downarrow$ & 0.038 & 0.036 & 0.033 & 0.036 & 0.030 & 0.019 & 0.029 & 0.024 & 0.027 & 0.029 & 0.032 \\ 
    \hline
  \multirow{4}{*}{I./ChineseAd} 
    & $S_{\alpha}~\uparrow$ & 0.676 & 0.701 & 0.670 & 0.655 & 0.801 & 0.822 & 0.732 & 0.658 & 0.743 & 0.670 & 0.702 \\
    & max $E_{\phi}~\uparrow$ & 0.977 & 0.972 & 0.951 & 0.922 & 0.980 & 0.988 & 0.969 & 0.949 & 0.967 & 0.942 & 0.979 \\
    & mean $E_{\phi}~\uparrow$ & 0.811 & 0.722 & 0.641 & 0.774 & 0.954 & 0.948 & 0.773 & 0.773 & 0.887 & 0.838 & 0.679 \\
    & $\mathcal{M}~\downarrow$ & 0.010 & 0.004 & 0.014 & 0.005 & 0.003 & 0.003 & 0.005 & 0.012 & 0.006 & 0.006 & 0.009\\ 
    \hline
  \multirow{4}{*}{I./Spanish} 
    & $S_{\alpha}~\uparrow$ & 0.773 & 0.808 & 0.884 & 0.827 & 0.844 & 0.868 & 0.830 & 0.842 & 0.862 & 0.812 & 0.740 \\
    & max $E_{\phi}~\uparrow$ & 0.929 & 0.963 & 0.969 & 0.968 & 0.969 & 0.974 & 0.964 & 0.956 & 0.964 & 0.928 & 0.963 \\
    & mean $E_{\phi}~\uparrow$ & 0.781 & 0.887 & 0.961 & 0.948 & 0.931 & 0.947 & 0.891 & 0.912 & 0.935 & 0.863 & 0.654 \\
    & $\mathcal{M}~\downarrow$ & 0.030 & 0.029 & 0.017 & 0.023 & 0.021 & 0.020 & 0.026 & 0.020 & 0.022 & 0.026 & 0.038 \\ 
    \hline
  \multirow{4}{*}{I./PianoSaxophone} 
    & $S_{\alpha}~\uparrow$ & 0.729 & 0.712 & 0.775 & 0.779 & 0.854 & 0.879 & 0.812 & 0.836 & 0.896 & 0.861 & 0.788 \\
    & max $E_{\phi}~\uparrow$ & 0.980 & 0.984 & 0.983 & 0.985 & 0.990 & 0.992 & 0.988 & 0.957 & 0.990 & 0.989 & 0.990 \\
    & mean $E_{\phi}~\uparrow$ & 0.732 & 0.663 & 0.951 & 0.888 & 0.956 & 0.932 & 0.858 & 0.942 & 0.936 & 0.914 & 0.703 \\
    & $\mathcal{M}~\downarrow$ & 0.013 & 0.014 & 0.015 & 0.011 & 0.010 & 0.008 & 0.012 & 0.012 & 0.009 & 0.009 & 0.016 \\ 
    \hline
  \multirow{4}{*}{I./BadmintonConvo} 
    & $S_{\alpha}~\uparrow$ & 0.572 & 0.600 & 0.765 & 0.648 & 0.693 & 0.909 & 0.694 & 0.648 & 0.771 & 0.620 & 0.610 \\
    & max $E_{\phi}~\uparrow$ & 0.775 & 0.937 & 0.928 & 0.875 & 0.913 & 0.977 & 0.949 & 0.751 & 0.941 & 0.718 & 0.965 \\
    & mean $E_{\phi}~\uparrow$ & 0.562 & 0.590 & 0.866 & 0.712 & 0.776 & 0.941 & 0.734 & 0.669 & 0.839 & 0.622 & 0.566  \\
    & $\mathcal{M}~\downarrow$ & 0.105 & 0.102 & 0.065 & 0.087 & 0.078 & 0.033 & 0.081 & 0.090 & 0.065 & 0.098 & 0.103 \\ 
    \hline
  \multirow{4}{*}{O./Beach} 
    & $S_{\alpha}~\uparrow$ & 0.615 & 0.656 & 0.607 & 0.688 & 0.693 & 0.636 & 0.648 & 0.668 & 0.632 & 0.668 & 0.593 \\
    & max $E_{\phi}~\uparrow$ & 0.858 & 0.906 & 0.866 & 0.907 & 0.832 & 0.925 & 0.932 & 0.822 & 0.928 & 0.929 & 0.940 \\
    & mean $E_{\phi}~\uparrow$ & 0.687 & 0.689 & 0.572 & 0.791 & 0.808 & 0.693 & 0.769 & 0.612 & 0.644 & 0.779 & 0.677 \\
    & $\mathcal{M}~\downarrow$ & 0.002 & 0.002 & 0.003 & 0.001 & 0.001 & 0.002 & 0.002 & 0.003 & 0.003 & 0.002 & 0.005 \\ 
    \hline
  \multirow{4}{*}{I./Brothers} 
    & $S_{\alpha}~\uparrow$ & 0.697 & 0.733 & 0.793 & 0.733 & 0.766 & 0.839 & 0.771 & 0.812 & 0.823 & 0.820 & 0.792 \\
    & max $E_{\phi}~\uparrow$ & 0.946 & 0.952 & 0.961 & 0.955 & 0.964 & 0.975 & 0.970 & 0.954 & 0.966 & 0.961 & 0.975 \\
    & mean $E_{\phi}~\uparrow$ & 0.732 & 0.761 & 0.945 & 0.875 & 0.904 & 0.929 & 0.846 & 0.932 & 0.909 & 0.922 & 0.708 \\
    & $\mathcal{M}~\downarrow$ & 0.019 & 0.017 & 0.015 & 0.015 & 0.015 & 0.012 & 0.017 & 0.012 & 0.015 & 0.013 & 0.020 \\ 
    \hline
  \multirow{4}{*}{I./Blues} 
    & $S_{\alpha}~\uparrow$ & 0.864 & 0.841 & 0.840 & 0.851 & 0.860 & 0.887 & 0.889 & 0.819 & 0.857 & 0.865 & 0.786 \\
    & max $E_{\phi}~\uparrow$ & 0.989 & 0.989 & 0.985 & 0.989 & 0.992 & 0.990 & 0.993 & 0.984 & 0.994 & 0.984 & 0.993 \\
    & mean $E_{\phi}~\uparrow$ & 0.922 & 0.897 & 0.964 & 0.970 & 0.971 & 0.964 & 0.935 & 0.916 & 0.907 & 0.954 & 0.688 \\
    & $\mathcal{M}~\downarrow$ & 0.012 & 0.011 & 0.011 & 0.010 & 0.008 & 0.007 & 0.010 & 0.008 & 0.011 & 0.009 & 0.015 \\ 
    \hline
  \multirow{4}{*}{I./Questions} 
    & $S_{\alpha}~\uparrow$ & 0.640 & 0.713 & 0.705 & 0.655 & 0.721 & 0.808 & 0.699 & 0.858 & 0.808 & 0.738 & 0.768 \\
    & max $E_{\phi}~\uparrow$ & 0.987 & 0.990 & 0.986 & 0.988 & 0.990 & 0.993 & 0.990 & 0.994 & 0.992 & 0.988 & 0.995 \\
    & mean $E_{\phi}~\uparrow$ & 0.630 & 0.774 & 0.852 & 0.761 & 0.883 & 0.887 & 0.725 & 0.958 & 0.896 & 0.782 & 0.656 \\
    & $\mathcal{M}~\downarrow$ & 0.015 & 0.014 & 0.013 & 0.014 & 0.011 & 0.009 & 0.012 & 0.008 & 0.010 & 0.011 & 0.016 \\ 
    \hline
  \multirow{4}{*}{O./Tennis} 
    & $S_{\alpha}~\uparrow$ & 0.762 & 0.689 & 0.825 & 0.809 & 0.826 & 0.847 & 0.816 & 0.855 & 0.807 & 0.815 & 0.782 \\
    & max $E_{\phi}~\uparrow$ & 0.967 & 0.967 & 0.985 & 0.976 & 0.979 & 0.984 & 0.983 & 0.980 & 0.979 & 0.978 & 0.988 \\
    & mean $E_{\phi}~\uparrow$ & 0.791 & 0.644 & 0.938 & 0.918 & 0.955 & 0.917 & 0.867 & 0.959 & 0.908 & 0.867 & 0.696 \\
    & $\mathcal{M}~\downarrow$ & 0.014 & 0.013 & 0.009 & 0.009 & 0.009 & 0.009 & 0.010 & 0.009 & 0.010 & 0.010 & 0.017 \\ 
    \hline
  \multirow{4}{*}{O./Sonwfield} 
    & $S_{\alpha}~\uparrow$ & 0.883 & 0.882 & 0.944 & 0.937 & 0.917 & 0.949 & 0.917 & 0.920 & 0.943 & 0.960 & 0.862 \\
    & max $E_{\phi}~\uparrow$ & 0.981 & 0.989 & 0.994 & 0.989 & 0.988 & 0.993 & 0.989 & 0.977 & 0.993 & 0.994 & 0.994 \\
    & mean $E_{\phi}~\uparrow$ & 0.900 & 0.906 & 0.987 & 0.974 & 0.973 & 0.969 & 0.937 & 0.932 & 0.974 & 0.980 & 0.744 \\
    & $\mathcal{M}~\downarrow$ & 0.023 & 0.020 & 0.008 & 0.013 & 0.014 & 0.011 & 0.017 & 0.017 & 0.011 & 0.009 & 0.032  \\ 
    \hline
  \multirow{4}{*}{I./PianoMono} 
    & $S_{\alpha}~\uparrow$ & 0.751 & 0.732 & 0.875 & 0.760 & 0.841 & 0.892 & 0.853 & 0.816 & 0.868 & 0.785 & 0.842 \\
    & max $E_{\phi}~\uparrow$ & 0.975 & 0.978 & 0.981 & 0.966 & 0.977 & 0.985 & 0.981 & 0.963 & 0.980 & 0.965 & 0.990  \\
    & mean $E_{\phi}~\uparrow$ & 0.776 & 0.749 & 0.967 & 0.923 & 0.938 & 0.965 & 0.923 & 0.944 & 0.932 & 0.836 & 0.752 \\
    & $\mathcal{M}~\downarrow$ & 0.034 & 0.034 & 0.021 & 0.037 & 0.025 & 0.020 & 0.026 & 0.036 & 0.026 & 0.028 & 0.031  \\ 
  \bottomrule
  \end{tabular}
  }
  \caption{Sequence-specific performance comparison of 8/2/1 state-of-the-art SOD/VSOD/360° SOD methods. I. = indoor. O. = outdoor.}
   \label{tab:seq_results}
\end{table*}

\begin{table*}[t!]
   \small
  \centering
  \renewcommand{\arraystretch}{0.3}
  \setlength\tabcolsep{2pt}
  \resizebox{0.99\textwidth}{!}{
  \begin{tabular}{cl||cccc|cc|c}
   \toprule
   \multicolumn{2}{l||}{\multirow{2}{*}{{Sequence}}}& \multicolumn{4}{c|}{General} & \multicolumn{2}{c|}{360°} & \multirow{2}{*}{{No.}}\\
   \cline{3-8}\\
   & & Multiple Persons & Occlusions & Distant View & Motion Blur & Out-of-View & Geometric Distortion
   \\
 \midrule
  \multirow{58}{*}{\begin{sideways}Indoor\end{sideways}} &
  French & & \ding{52} & & \ding{52} & \ding{52} & \ding{52} & 4  \\
  & WaitingRoom & \ding{52} & \ding{52} & \ding{52} & \ding{52} & & & 4 \\
  & Cooking & \ding{52} & \ding{52} & \ding{52} & \ding{52} & & \ding{52} & 5 \\
  & Guitar & \ding{52} & & \ding{52} & & & \ding{52} & 3 \\
  & Warehouse & & & \ding{52} & \ding{52} & & \ding{52}& 3 \\
  & Bass & \ding{52} & \ding{52} & \ding{52} & & & \ding{52} & 4 \\
  & Passageway & & \ding{52} & \ding{52} & \ding{52} & \ding{52} & \ding{52} & 5 \\
  & RuralDriving & & & & & & \ding{52} & 1  \\
  & MICOSinging & & \ding{52} & & & & \ding{52} & 2 \\
  & ScenePlay & \ding{52} & & & & & \ding{52} & 2 \\
  & UrbanDriving & & & & & & \ding{52} & 1 \\
  & Clarinet & \ding{52} & \ding{52} & & & & \ding{52} & 3  \\
  & Interview & &\ding{52} & \ding{52} & & & & 2  \\
  & Telephone & & & & \ding{52} & & \ding{52} & 2 \\
  & Breakfast & & \ding{52} & \ding{52} & \ding{52} & & \ding{52} & 4 \\
  & PianoSaxophone & \ding{52} & \ding{52} & \ding{52} & & \ding{52} & \ding{52} & 5 \\
  & Chorus & \ding{52} & \ding{52} & \ding{52} & & & \ding{52} & 4  \\
  & Studio & \ding{52} & & \ding{52} & \ding{52} & & \ding{52} & 4 \\
  & BadmintonConvo & \ding{52} & \ding{52} & & \ding{52} & \ding{52} & \ding{52} & 5 \\
  & Director & \ding{52} & \ding{52} & \ding{52} & \ding{52} & & \ding{52} & 5 \\
  & ChineseAd & &\ding{52} & \ding{52} & \ding{52} & \ding{52} & & 4 \\
  & Blues & \ding{52} & \ding{52} & \ding{52} & & & \ding{52} & 4 \\
  & Brothers & \ding{52} & \ding{52} & \ding{52} & \ding{52} & \ding{52} & \ding{52} & 6 \\
  & Violins & \ding{52} & & \ding{52} & & \ding{52} & \ding{52} & 4 \\
  & Spanish & & \ding{52} & & \ding{52} & \ding{52} & \ding{52} & 4  \\
  & Questions & \ding{52} & \ding{52} & \ding{52} & \ding{52} & & \ding{52} & 5 \\
  & PianoMono & & & & \ding{52} & & \ding{52} & 2 \\
  & Melodrama & \ding{52} & & & & & \ding{52} & 2 \\
  & SingingDancing & \ding{52} & \ding{52} & \ding{52} & \ding{52} & & \ding{52} & 5\\
  \midrule
  \multirow{24}{*}{\begin{sideways}Outdoor\end{sideways}} &
  Bicycling & & \ding{52} & \ding{52} & \ding{52} & \ding{52} & \ding{52} & 5  \\
  & Japanese & & & & \ding{52} & \ding{52} & \ding{52} & 3 \\
  & Surfing & & \ding{52} & & & & & 1 \\
  & Lawn & & & \ding{52} & \ding{52} & & \ding{52} & 3 \\
  & AudiAd & \ding{52}  & \ding{52} & \ding{52} & \ding{52} & & & 4 \\
  & Walking & & & \ding{52} & \ding{52} & & \ding{52} & 3 \\
  & Bridge & & \ding{52} & \ding{52} & \ding{52} & & & 3 \\
  & RacingCar & & & & & & \ding{52} & 1 \\
  & Football & & & \ding{52} & \ding{52} & \ding{52} & & 3 \\
  & Snowfield & & & & \ding{52} &  & \ding{52} & 2 \\
  & Beach & \ding{52} & \ding{52} & \ding{52} & \ding{52} & & & 4 \\
  & Tennis & \ding{52} & \ding{52} & \ding{52} & \ding{52} & \ding{52} & \ding{52} & 6 \\
  \midrule
  \\
  No. & & 20 & 25 & 26 & 26 & 12 & 33 & 142 \\
  \bottomrule
  \end{tabular}}
   \caption{Per-sequence attribute statistics.}
   \label{tab:attributes_details}
\end{table*}

\begin{figure*}[t!]
	\centering
	\begin{overpic}[width=0.99\textwidth]{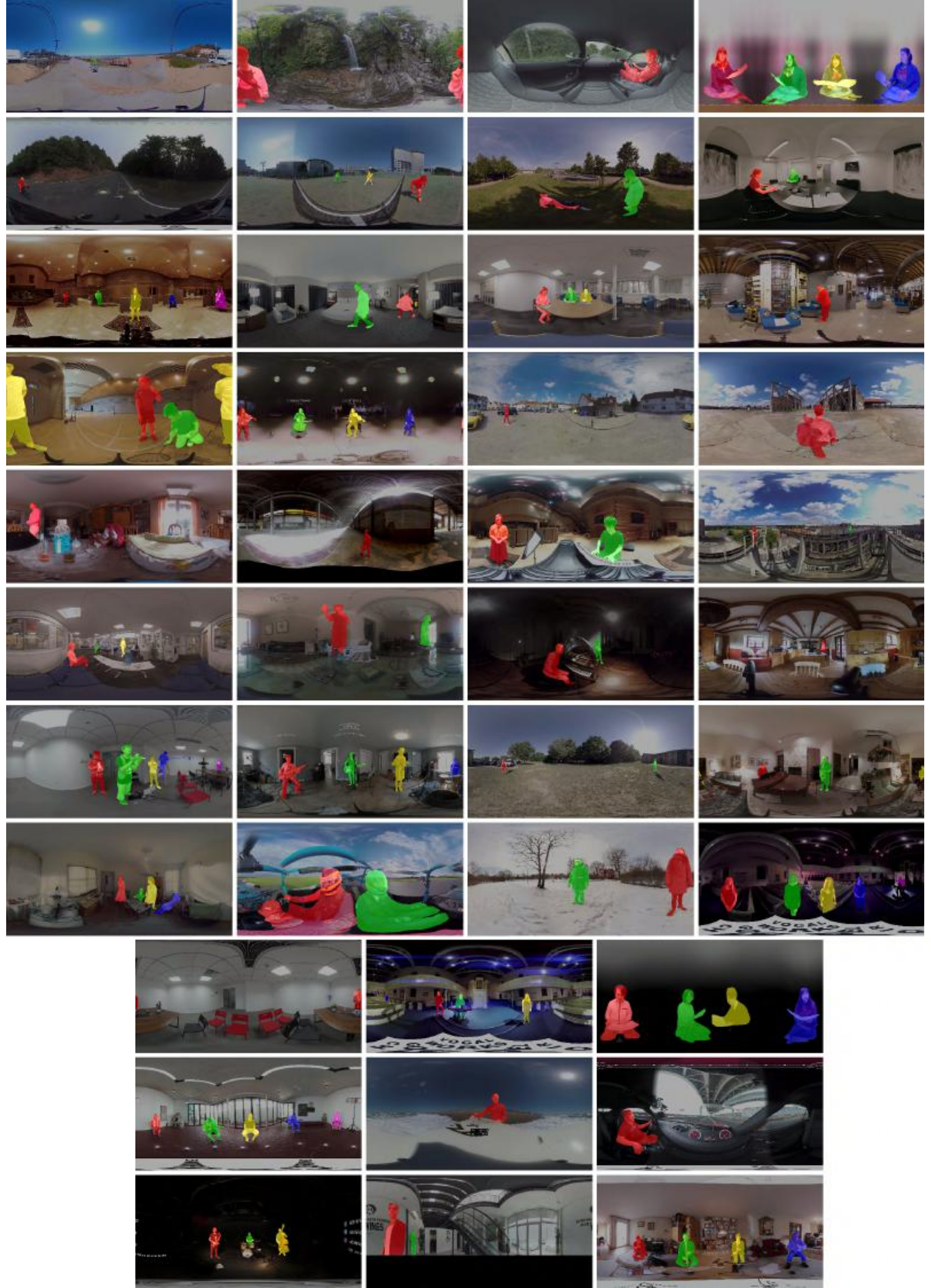}
    \end{overpic}
	\caption{Sample frames from \ourdataset, with instance-level ground truth overlaid.}
    \label{fig:annotation_sum}
\end{figure*}

\end{appendices}

\clearpage
\bibliographystyle{neurips_data_2021}
\bibliography{neurips_data_2021}

\end{document}